\documentclass{article}

\PassOptionsToPackage{numbers, compress}{natbib}
 \usepackage[preprint]{neurips_2026}

\usepackage[utf8]{inputenc} 
\usepackage[T1]{fontenc}    
\usepackage{hyperref}       
\usepackage{url}            
\usepackage{booktabs}       
\usepackage{amsfonts}       
\usepackage{nicefrac}       
\usepackage{microtype}      
\usepackage{xcolor}         

\usepackage{graphicx}
\usepackage{tabularx}
\usepackage{amsmath}

\usepackage{array}
\usepackage{ragged2e}

\newcolumntype{L}[1]{>{\RaggedRight\arraybackslash}p{#1}}
\newcolumntype{Y}{>{\RaggedRight\arraybackslash}X}

\usepackage{chngcntr}
\usepackage{longtable}

\title{A Scientific Claim Stability Framework for Evaluating~MRI Morphometry Pipelines}

\author{%
Duy-Cat Can\textsuperscript{1,2,3},
Dang-Duat Tran\textsuperscript{3},
Anh-Khang Nguyen\textsuperscript{3},
Quoc-Bao Tran\textsuperscript{3},
Trung-Hieu Do\textsuperscript{4,5},\\
\textbf{%
Van-Tu Nguyen\textsuperscript{6},
Duy-Thanh Vu\textsuperscript{1,2,3},
Oliver Y. Ch\'en\textsuperscript{1,2,\S},
and Hoang-Quynh Le\textsuperscript{3,\S}
}\vspace{.5em}
\\
for the Alzheimer's Disease Neuroimaging Initiative\thanks{\tiny Part of data used in preparation of this article were obtained from the Alzheimer's Disease Neuroimaging Initiative (ADNI) database (\href{https://adni.loni.usc.edu/}{adni.loni.usc.edu}). As such, the investigators within the ADNI contributed to the design and implementation of ADNI and/or provided data but did not participate in the analysis or writing of this report. A complete listing of ADNI investigators can be found at: \href{https://adni.loni.usc.edu/wp-content/uploads/how_to_apply/ADNI_Acknowledgement_List.pdf}{ADNI Acknowledgment List}}\\
and the Australian Imaging Biomarkers and Lifestyle flagship study of aging\thanks{\tiny Part of data used in the preparation of this article was obtained from the Australian Imaging Biomarkers and Lifestyle flagship study of aging (AIBL) funded by the Commonwealth Scientific and Industrial Research Organization (CSIRO) which was made available at the ADNI database (\href{https://adni.loni.usc.edu/}{adni.loni.usc.edu}). The AIBL researchers contributed data but did not participate in analysis or writing of this report. AIBL researchers are listed at \href{https://data.aibl.org.au/adni/}{data.aibl.org.au/adni/}.}\\
[0.7em]
\textsuperscript{1}Centre for Neural Artificial Intelligence, Lausanne University Hospital, Lausanne, Switzerland\\
\textsuperscript{2}Faculty of Biology and Medicine, University of Lausanne, Lausanne, Switzerland\\
\textsuperscript{3}VNU University of Engineering and Technology, Hanoi, Vietnam\\
\textsuperscript{4}Hanoi Medical University, Hanoi, Vietnam\\
\textsuperscript{5}National Geriatric Hospital, Hanoi, Vietnam\\
\textsuperscript{6}Institute of Diagnostic and Interventional Radiology, Bach Mai Hospital, Hanoi, Vietnam\\[0.5em]
\texttt{duy-cat.can@chuv.ch, olivery.chen@chuv.ch}\\
\texttt{lhquynh@vnu.edu.vn}\\
\textsuperscript{\S}Corresponding authors
}

\begin{document}

\maketitle

\begin{abstract}
MRI morphometry pipelines are often evaluated by measurement agreement or downstream accuracy, yet their scientific value depends on the conclusions they support.
Two pipelines can produce similar feature-level results yet support different conclusions about aging, cognitive decline, or neurodegeneration. We introduce Scientific Claim Stability (SCS), a claim-centric evaluation framework that converts region-of-interest (ROI) morphometry tables into auditable claim cards and evaluates their support under changes in dataset, processing pipeline, atlas, feature family, and target.
Across 10{,}370 MRI scans from three datasets and five cohorts, SCS generates 2{,}783 claim cards and evaluates 89 claim-family pairs. The evaluation distinguishes stable, transferable, and fragile claims. FreeSurfer 8--FastSurfer comparisons show high evidence concordance and low anatomical-support overlap under the primary covariate specification. Acquisition-adjusted analyses show that the magnitude of support disagreement depends on the model specification.
Structured review by specialists in geriatrics, neuroscience, and radiology, alongside auxiliary ChatGPT and Gemini reviewers, assesses claim plausibility, clarity, faithfulness, usefulness, and caution. SCS provides a reproducible way to audit the evidential stability of morphometry claims without assuming a universal gold standard.
\end{abstract}

\section{Introduction}
\label{sec:introduction}

Structural MRI morphometry is widely used to support scientific claims about aging, cognitive decline, and neurodegeneration.
Large datasets such as ADNI~\citep{petersen2010alzheimers}, OASIS-1~\citep{marcus2007open}, and AIBL~\citep{ellis2009australian} have made region-of-interest (ROI) morphometry widely available for biomarker discovery and machine learning.
Yet the same images can support different conclusions depending on the processing pipeline, feature family, atlas, covariate model, preprocessing choice, and cohort definition.
This sensitivity is well documented: FreeSurfer version changes can alter morphometric measurements~\citep{gronenschild2012effects}, alternative pipelines encode different assumptions~\citep{fischl2012freesurfer,henschel2020fastsurfer,ashburner2000voxel}, and broader analytic flexibility can change scientific inferences even when individual choices are defensible~\citep{carp2012plurality,simonsohn2020specification}.

Existing evaluations assess morphometric measurement agreement, segmentation accuracy, test--retest reliability, and downstream predictive performance~\citep{fischl2012freesurfer,henschel2020fastsurfer,iscan2015test,knussmann2022test}.
These assessments characterize pipeline outputs, while scientific conclusions also depend on which regional associations support each statement.
Two pipelines may have high ROI-level evidence concordance while selecting different brain regions to support a diagnosis, cognition, or age-related statement.
The remaining question is whether the same scientific claim remains supported, transferable, and interpretable under realistic dataset, pipeline, atlas, feature-family, and target perturbations.

We introduce Scientific Claim Stability (SCS), a claim-centric framework for evaluating MRI morphometry pipelines. SCS uses scientific claims as its unit of evaluation.
It converts ROI morphometry tables into structured claim cards that link a natural-language statement to evidence, support regions, weights, provenance metadata, stability and transfer profiles, and review status.
For example, a card describing an association between cortical thickness in a named brain region and MMSE records the supporting statistics and analysis setting.
Claim cards extend the structured documentation used in model cards and datasheets~\citep{mitchell2019model,gebru2021datasheets} to scientific conclusions.
We implement \texttt{scs\_morph}, a configuration-driven Python package for claim extraction, cross-setting evaluation, and expert review.

We evaluate SCS on 10{,}370 MRI scans from ADNI, OASIS-1, and AIBL.
The framework generates~121 claim families and 2{,}783 claim cards, then evaluates 89 claim-family pairs across dataset transfer, weak-label transfer, pipeline drift, atlas sensitivity, and feature-family sensitivity.
The audit identifies a central failure mode: \emph{high evidence concordance does not guarantee scientific claim stability}.
In the FreeSurfer--FastSurfer volume comparison under the primary covariate specification, evidence vectors remained highly concordant while supporting regions had very low overlap, making all eight evaluated pairs fragile.
Additional acquisition-adjusted analyses show that the magnitude of support disagreement depends on the analysis specification.
Claim-level evaluation thus identifies differences in reliability that evidence agreement alone can miss.

\begin{figure}[t]
\centering
\includegraphics[width=\textwidth]{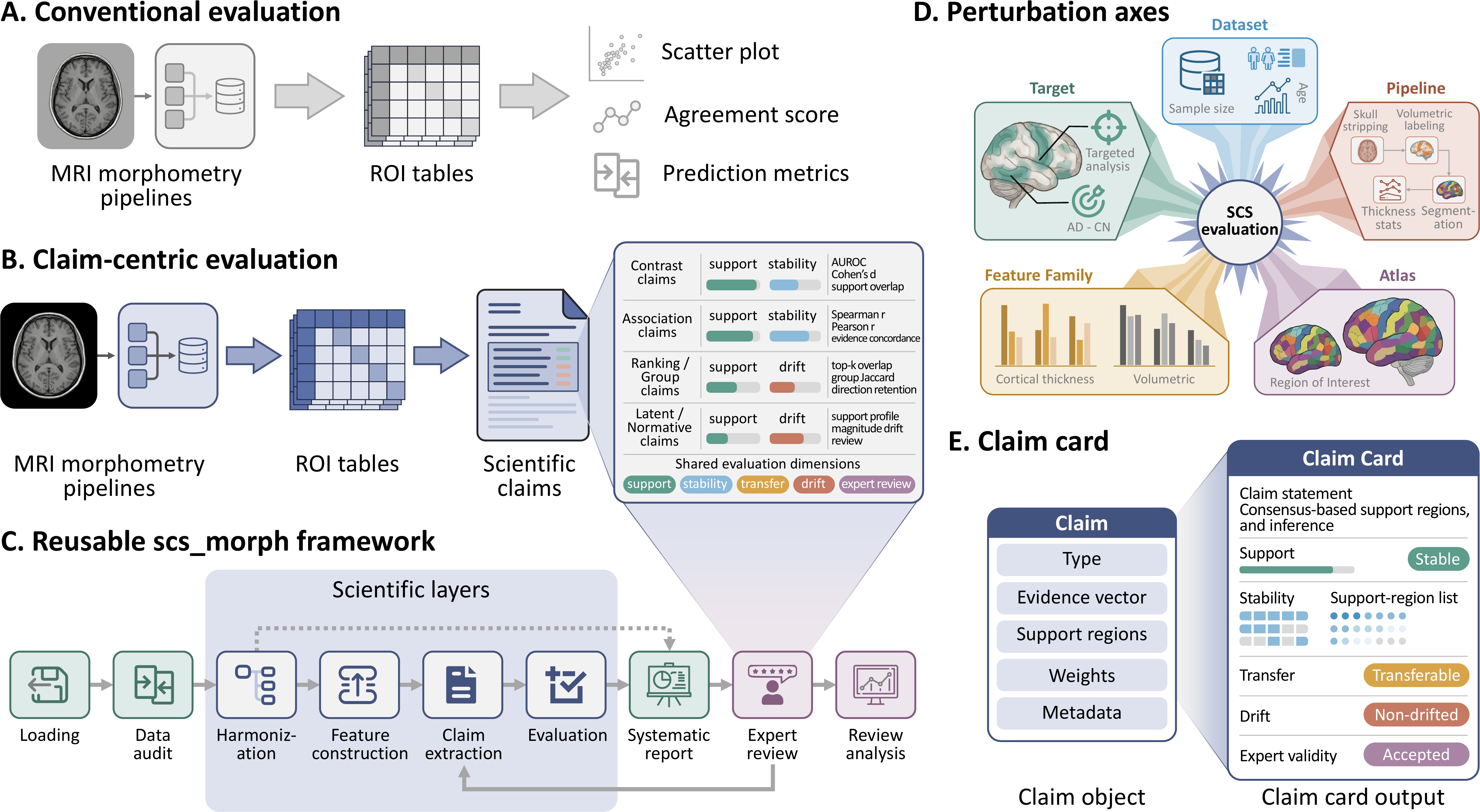}
\caption{
Overview of the Scientific Claim Stability framework.
Conventional evaluation compares ROI measurements or prediction scores, whereas SCS converts ROI tables into claim cards and evaluates whether the resulting scientific conclusions remain supported under dataset, pipeline, atlas, feature-family, and target perturbations.
Details in Appendix~\ref{app:figure_scs_framework}.
}
\label{fig:scs_framework}
\end{figure}

This paper makes five contributions.
\textbf{First}, we formalize scientific claims as typed, transferable evaluation objects extracted from ROI morphometry tables.
\textbf{Second}, we introduce claim cards and multidimensional stability profiles for documenting interpretable morphometry conclusions.
\textbf{Third}, we provide \texttt{scs\_morph}, a reusable framework for claim extraction, evaluation, and expert/LLM review.
\textbf{Fourth}, we audit 2{,}783 generated claim cards and 89 evaluated claim-family pairs across datasets, pipelines, atlases, feature families, and targets.
\textbf{Fifth}, we evaluate selected generated claims through structured human expert and LLM review.
Figure~\ref{fig:scs_framework} summarizes the proposed shift from measurement-level evaluation to claim-level evaluation.

\section{Related Work and Evaluation Gap}
\label{sec:related_work}

\textbf{MRI morphometry pipelines and drift.}
Structural MRI morphometry pipelines convert anatomical scans into quantitative features such as cortical thickness, cortical regional volume, and subcortical volume.
FreeSurfer is widely used for surface-based morphometry~\citep{fischl2012freesurfer}, while SPM/CAT-style workflows provide alternative voxel-based and morphometric analyses~\citep{ashburner2000voxel,dias2022cortical}.
FastSurfer and SynthSeg further extend the ecosystem with deep-learning-based processing designed for speed or robustness across imaging conditions~\citep{henschel2020fastsurfer,billot2023synthseg}.
However, prior work shows that morphometry estimates can vary across software versions, computing environments, sites, scanners, and quality-control procedures~\citep{gronenschild2012effects,iscan2015test,knussmann2022test}.
FreeSurfer version changes can affect regional measures and case-control statistical outcomes~\citep{filip2022different}, and FreeSurfer--CAT12 comparisons show that high within-pipeline reliability does not guarantee strong between-pipeline reproducibility at the ROI level~\citep{dias2022cortical,schmitzkoep2025longitudinal}.
These studies establish the need for pipeline evaluation through analyses of measurements, reliability, and compatibility; SCS extends their scope to the stability of the claims supported by those measurements.

\textbf{Analytic flexibility and partially observed multiverses.}
Neuroimaging results depend on a range of defensible analytic choices~\citep{carp2012plurality}.
Specification-curve and multiverse analyses make this dependence explicit by examining how conclusions change across reasonable specifications~\citep{simonsohn2020specification,steegen2016increasing}.
Recent work on multiverse sampling addresses the difficulty of exhaustive enumeration when the space of possible pipelines is large~\citep{short2025lost}.
SCS adopts this perspective but adapts it to a practical morphometry setting in which the multiverse is only partially observed: not every dataset is available under every pipeline, atlas, feature family, target, or covariate specification.
SCS identifies stable, transferable, and fragile claims and quantifies their stability under the perturbations that can actually be evaluated.

\textbf{Documentation artifacts and human-centered evaluation.}
Model cards and datasheets established structured documentation for model behavior and dataset provenance~\citep{mitchell2019model,gebru2021datasheets}.
Recent work extends this idea to dataset and benchmark documentation, including Croissant for ML-ready dataset metadata and BenchmarkCards for benchmark transparency~\citep{akhtar2024croissant,sokol2025benchmarkcards}.
SPHERE similarly frames evaluation cards as a way to document human-AI evaluation design and validity~\citep{zhao2025sphere}.
SCS extends this documentation approach to scientific conclusions.
Each claim card links a conclusion to evidence, supporting brain regions, perturbation stability, transfer behavior, and expert or LLM review.
The scientific claim itself becomes the object of documentation and evaluation.

\section{Scientific Claim Stability Framework}
\label{sec:framework}

\subsection{The observed morphometry multiverse}
\label{sec:observed_multiverse}

A morphometry conclusion depends on the complete analysis specification, including the dataset.
We represent one specification as
\begin{equation}
    s \in \mathcal{S}
    =
        \mathcal{D}
        \times
        \mathcal{H}
        \times
        \mathcal{P}
        \times
        \mathcal{F}
        \times
        \mathcal{A}
        \times
        \mathcal{T}
        \times
        \mathcal{C},
    \label{eq:analysis_setting_space}
\end{equation}
where $\mathcal{D}$ indexes datasets or views, $\mathcal{H}$ cohorts or subsets, $\mathcal{P}$ pipelines, $\mathcal{F}$ feature families, $\mathcal{A}$ atlases or region systems, $\mathcal{T}$ scientific targets, and $\mathcal{C}$ covariate specifications.
These choices jointly define the evaluation object, consistent with the dependence of scientific conclusions on analytic specifications~\citep{carp2012plurality,simonsohn2020specification}.

Each setting $s$ induces an aligned analysis table
\begin{equation}
\Psi(s)
=
(\mathbf{X}_s,\mathbf{Z}_s,\mathbf{y}_s),
\qquad
\mathbf{X}_s\in\mathbb{R}^{n_s\times p_s},
\quad
\mathbf{Z}_s\in\mathbb{R}^{n_s\times q_s},
\label{eq:analysis_table}
\end{equation}
where $\mathbf{X}_s$ is the scan-by-region morphometry matrix, $\mathbf{Z}_s$ contains metadata and covariates, and $\mathbf{y}_s$ is an optional target such as diagnosis, MMSE, age, acquisition group, or CDR-derived impairment.
The ideal multiverse $\mathcal{M}$ contains all defensible settings, whereas the observed multiverse
\begin{equation}
\mathcal{M}_{\mathrm{obs}}
=
\{s_1,\ldots,s_m\}
\subset
\mathcal{M}
\subseteq
\mathcal{S}
\label{eq:observed_multiverse}
\end{equation}
contains only settings that are computationally, institutionally, and empirically available.
SCS evaluates stability under the perturbations available in this partially observed multiverse and records unavailable or insufficient comparisons.
Operationally, the observed multiverse is a graph: each node is an available analysis setting, and each evaluated edge connects settings through a declared dataset, pipeline, atlas, feature-family, or target change.
Here, 121 claim-family settings define 89 predefined evaluation edges, with inference restricted to observed combinations.
Appendix~\ref{app:settings} gives the full setting schema, feature-space expansion rules, and matching definitions.

\subsection{Claim cards and extraction}
\label{sec:claim_objects}

SCS defines the scientific claim as the unit of evaluation.
For a setting $s\in\mathcal{M}_{\mathrm{obs}}$, a claim card is
\begin{equation}
c_s
=
(\tau,q,\mathbf{e},S_k,\mathbf{w},M),
\label{eq:claim_card}
\end{equation}
where $\tau$ is the claim type, $q$ is the natural-language statement, $\mathbf{e}\in\mathbb{R}^{p_s}$ is a signed evidence vector, $S_k\subseteq\{1,\ldots,p_s\}$ is the support set, $\mathbf{w}\in\mathbb{R}^{p_s}$ is a claim weight vector, and $M$ stores provenance, covariates, feature-space metadata, extraction criteria, and status information.
The human view of the card is $\mathrm{view}_{\mathrm{human}}(c_s)=q$.
The machine view is $\mathrm{view}_{\mathrm{machine}}(c_s)=(\mathbf{e},S_k,\mathbf{w},M)$.
The two views link each statement to its evidence, support regions, and evaluation profile.

Claim cards are generated by a typed evidence operator,
\begin{equation}
\mathcal{E}_{\tau}:
(\mathbf{X}_s,\mathbf{y}_s,\mathbf{Z}_s)
\mapsto
c_s.
\label{eq:claim_extraction_operator}
\end{equation}
For supervised claim types, SCS fits ROI-wise covariate-adjusted association models,
\begin{equation}
x_{ij}
=
\alpha_j
+
\beta_j y_i
+
\mathbf{z}_i^\top\boldsymbol{\gamma}_j
+
\epsilon_{ij},
\qquad
i=1,\ldots,n_s,
\qquad
j=1,\ldots,p_s,
\label{eq:roiwise_model}
\end{equation}
where $\epsilon_{ij}$ denotes the residual error for scan $i$ and region $j$.
The fitted target association is converted into signed regional evidence 
%
%
$e_j=T(\hat{\beta}_j)$,
where $T$ may return a signed coefficient, standardized coefficient, effect size, or $t$-statistic.
The support set $S_k(c_s)$ contains regions with the strongest evidence and is checked against explicit support criteria.
A simplified support indicator is
\begin{equation}
\mathbb{I}_{\mathrm{support}}(c_s)
=
\mathbb{I}
\left[
|S_k(c_s)|\geq k_{\min},
\quad
\overline{|e|}_{S_k}\geq \eta,
\quad
n_s\geq n_{\min}
\right],
\label{eq:support_indicator}
\end{equation}
where $\overline{|e|}_{S_k}$ is the mean absolute evidence over the support set.
The implementation also supports FDR thresholds, group-level support, direction consistency, and claim-level status assignment.
The same claim-card abstraction covers latent-axis and normative-deviation claims, where $\mathbf{e}$ is derived from unsupervised morphometric variation or deviation from a reference group.
Appendix~\ref{app:claim_schema} gives the full claim schema and taxonomy, and Appendix~\ref{app:roiwise_models} gives the statistical extraction details.

\subsection{Stability and transfer profiles}
\label{sec:claim_profiles}

SCS evaluates a source claim $c_s$ against a target setting $s_t$ through a multidimensional profile,
\begin{equation}
\mathcal{V}:(c_s,s_t)\mapsto\boldsymbol{\phi}(c_s,s_t),
\qquad
\boldsymbol{\phi}
=
(
\phi_{\mathrm{sup}},
\phi_{\mathrm{stab}},
\phi_{\mathrm{trans}},
\phi_{\mathrm{drift}},
\phi_{\mathrm{review}}
).
\label{eq:claim_profile}
\end{equation}
SCS reports each component of the profile separately.
A claim may be stable but not transferable, transferable but anatomically drifted, or fragile despite high evidence concordance.
The metrics are accompanied by descriptive labels such as stable, moderate, fragile, stable-and-transferable, stable-not-transferable, transferable-but-drifted, and insufficient.

For matched feature spaces, SCS separates evidence concordance (EC), support overlap (SO), and direction retention (DR):
\begin{equation}
\text{EC}(s,s_t)
=
\rho_{\mathrm{Spearman}}
\left(
\mathbf{e}_s[\mathcal{J}_{st}],
\mathbf{e}_t[\mathcal{J}_{st}]
\right),
\qquad
\text{SO}_k(s,s_t)
=
\frac{|S_k(c_s)\cap S_k(c_t)|}
{|S_k(c_s)\cup S_k(c_t)|},
\label{eq:ec_so}
\end{equation}
\begin{equation}
\text{DR}_k(s,s_t)
=
\frac{1}{|S_k(c_s)|}
\sum_{j\in S_k(c_s)}
\mathbb{I}
\left[
\operatorname{sign}(e_{s,j})
=
\operatorname{sign}(e_{t,j})
\right].
\label{eq:direction_retention}
\end{equation}
Here $\mathcal{J}_{st}$ is the matched ROI set, $\mathbf{e}_s$ and $\mathbf{e}_t$ are the source and target evidence vectors, $e_{s,j}$ and $e_{t,j}$ are their signed evidence values for region $j$, and $c_t$ is the corresponding target claim.
Evidence concordance measures whether the signed evidence ranking is preserved, support overlap measures whether the regions justifying the claim are preserved, and direction retention measures whether source-support regions retain their effect direction.
A high global evidence correlation can coexist with a fragile scientific conclusion when the support set or direction pattern changes.
Stability permits changes in effect magnitude when the evidence ranking, supporting regions, and effect directions remain sufficiently preserved.
The categorical thresholds are preset in the reported experiments and configurable in the software; the continuous metric profile remains the primary result (Appendix~\ref{app:labels}).

For non-identical feature spaces, SCS maps source and target regions to a shared group space,
\begin{equation}
g_s:\{1,\ldots,p_s\}\rightarrow\mathcal{G},
\qquad
g_t:\{1,\ldots,p_t\}\rightarrow\mathcal{G},
\label{eq:group_maps}
\end{equation}
and computes stability over shared groups such as lobes, network-like groups, ROI groups, or anatomical groups.
These mappings allow atlas and feature-family sensitivity to be evaluated in a shared group space.

Transfer is evaluated by applying the source claim as a fixed scoring functional in the target setting.
With a normalized source weight vector $\mathbf{w}_s$ restricted to source--target matched features, the transferred claim score for target scan $i$ is
\begin{equation}
u_i^{(t)}
=
\mathbf{x}_{t,i}^{\top}\mathbf{w}_s.
\label{eq:claim_transfer_score}
\end{equation}
Binary targets are evaluated with oriented AUROC and effect size (Cohen's $d$), while continuous targets are evaluated with correlation between $\mathbf{u}^{(t)}$ and $\mathbf{y}_t$.
Transfer tests whether a claim extracted in one setting remains useful in another with its weights held fixed.
Appendix~\ref{app:evaluation_metrics} gives exact matching, group matching, stability metrics, transfer metrics, and label thresholds.

\subsection{Expert and LLM review}
\label{sec:expert_llm_review}

SCS also evaluates the scientific usability of generated claims through structured review.
For each selected claim $c$, a human expert or auxiliary LLM reviewer $r$ assigns scores for plausibility, clarity, faithfulness, usefulness, and confidence (higher scores indicate better ratings or greater confidence):
\begin{equation}
a_{r,d}(c)
\in
\{1,\ldots,5\},
\qquad
d\in
\{
\mathrm{plausibility},
\mathrm{clarity},
\mathrm{faithfulness},
\mathrm{usefulness},
\mathrm{confidence}
\}.
\label{eq:review_scores}
\end{equation}
Reviewers also assign a caution level and recommended action: accept, revise, reject, or not sure.
The review profile summarizes mean scores, reviewer agreement, high-disagreement flags, corrections, and decisions.
LLM judges provide scalable auxiliary review, with agreement assessed against human expert scores.
The review outcomes are stored with the claim profile alongside statistical support, stability, and transfer measures.
Appendix~\ref{app:expert_review_protocol} gives the review prompt, scoring rubric, agreement metrics, and decision rules.

\section{Implementation and Empirical Design}
\label{sec:implementation_design}

\subsection{Implementation artifact}
\label{sec:implementation}

We implemented SCS as a configuration-driven Python framework, \texttt{scs\_morph}.
It converts harmonized MRI morphometry tables into claim cards and claim families, evaluates claim-family pairs across observed perturbations, and exports selected cards for expert and LLM review.
The workflow includes data audit, metadata harmonization, feature-matrix construction, claim extraction, stability and transfer evaluation, and review analysis.
It produces serialized claim cards, claim-family summaries, ROI-level statistics, evaluation-pair results, metric-specific summaries, and review tables.
The public implementation is available at \url{https://github.com/catcd/scs-morph}.
The repository contains the code, configurations, and scripts needed to reproduce the workflow after authorized access to the source datasets.
Appendix~\ref{app:implementation_details} describes the package structure, configuration files, output artifacts, and script-based reproducibility workflow.

\subsection{Datasets, targets, and feature spaces}
\label{sec:data_features}

We evaluate SCS on 10{,}370 MRI scans from the ADNI (\href{https://adni.loni.usc.edu}{adni.loni.usc.edu})~\citep{petersen2010alzheimers}, OASIS-1~\citep{marcus2007open}, and AIBL~\citep{ellis2009australian} datasets.
Established in 2003 as a public--private partnership, ADNI investigates whether serial MRI, PET, biological markers, and clinical and neuropsychological assessments can measure the progression of MCI and early AD.
Dataset-specific access and provenance are documented in Appendix~\ref{app:data_availability}.
The empirical multiverse contains full and baseline dataset views, plus an ADNI subset with matched FreeSurfer and SPM/CAT12 cortical-thickness features.
ADNI and AIBL are used with CN/MCI/AD-compatible labels, MMSE, age, sex, and acquisition metadata.
OASIS-1 uses CDR-derived cognitive impairment, MMSE, age, sex, and education, with a target definition distinct from the harmonized diagnosis labels.

The feature spaces span volumetric and cortical-thickness morphometry.
Volume features include FreeSurfer 8 cortical regional volumes and subcortical volumes~\citep{fischl2012freesurfer}.
Cortical-thickness features include FreeSurfer 8 Schaefer 200/7-network thickness~\citep{schaefer2018local}, Kong 200/17-network thickness~\citep{kong2022individualized}, and Yale Brain Atlas thickness~\citep{collins2025yale}.
Pipeline perturbation analyses use SPM/CAT12 Schaefer cortical-thickness features~\citep{ashburner2000voxel,dias2022cortical} and FastSurfer cortical and subcortical volumes~\citep{henschel2020fastsurfer}.
The ADNI SPM-matched view contains 2{,}892 scans with aligned FreeSurfer 8 and SPM/CAT12 Schaefer cortical-thickness matrices.
Appendix~\ref{app:datasets_features} gives the dataset-view inventory, feature-space inventory, matched matrix sizes, and data-access constraints.

\subsection{Claim generation and evaluation design}
\label{sec:claim_generation_evaluation_design}

Claims are generated at the analysis-setting level.
Each dataset--feature--target specification produces one claim family, and each claim family can contain multiple claim cards at different granularities.
Claim extraction produced 121 claim families and 2{,}783 claim cards.
Claim levels include composite, atomic ROI, group-summary, ranking, direction-summary, latent-axis, latent-axis association, normative-deviation, and normative atomic ROI claims.
Claim text is generated from structured templates, with each statement linked to source-setting metadata, ROI-level evidence, support regions, and support criteria.

We evaluated 89 claim-family pairs across six perturbation experiments.
\textbf{\texttt{E1}} evaluates ADNI-to-AIBL volume transfer.
\textbf{\texttt{E2}} evaluates ADNI-to-OASIS-1 volume transfer, including weak-label transfer from ADNI CN--AD claims to OASIS-1 CDR-derived impairment.
\textbf{\texttt{E3}} evaluates FreeSurfer 8 versus SPM/CAT12 cortical-thickness drift on the matched ADNI subset.
\textbf{\texttt{E4}} evaluates FreeSurfer 8 versus FastSurfer volume drift on identical ADNI baseline scans, target definitions, filters, and primary covariates (age, sex, education, and intracranial volume, as applicable).
\textbf{\texttt{E5}} evaluates FreeSurfer 8 cortical-thickness atlas sensitivity across Schaefer, Kong, and Yale representations.
\textbf{\texttt{E6}} evaluates FreeSurfer 8 volume-versus-thickness feature-family sensitivity.
Exact ROI matching is used when source and target settings share comparable feature names, and group-level matching is used when atlases or feature families do not admit one-to-one ROI correspondence.
All 89 configured evaluation pairs were evaluated with no skipped pairs and no processing errors.
Appendix~\ref{app:claim_generation_inventory} provides the claim-generation inventory, and Appendix~\ref{app:evaluation_experiment_details} gives the full \textbf{\texttt{E1}}--\textbf{\texttt{E6}} design and matching rules.

\subsection{Expert and LLM review design}
\label{sec:review_design}

The review workflow evaluates whether generated claim cards are plausible, clear, faithful to the displayed evidence, useful for interpretation, and appropriately cautious.
From the pool of 2{,}783 claim cards, the human expert review set contains 60 claims, all of which were also included in the 120-claim LLM review set along with 60 additional claims.
The three human reviewers, covering neuroscience, geriatrics, and radiology, each assessed the same 60 claims.
LLM judges provide auxiliary review, and their agreement with human experts is evaluated empirically.
Reviewers score plausibility, clarity, faithfulness, usefulness, and confidence on 1--5 scales, and also provide a caution level, recommended action, optional corrected claim text, and optional comments.
Appendix~\ref{app:expert_review_protocol} gives the full review protocol, prompt, scoring rubric, agreement metrics, and decision rules.

\section{Results}
\label{sec:results}

\subsection{SCS generates an auditable claim landscape}
\label{sec:results_claim_landscape}

\begin{figure}[t]
\centering
\includegraphics[width=\textwidth]{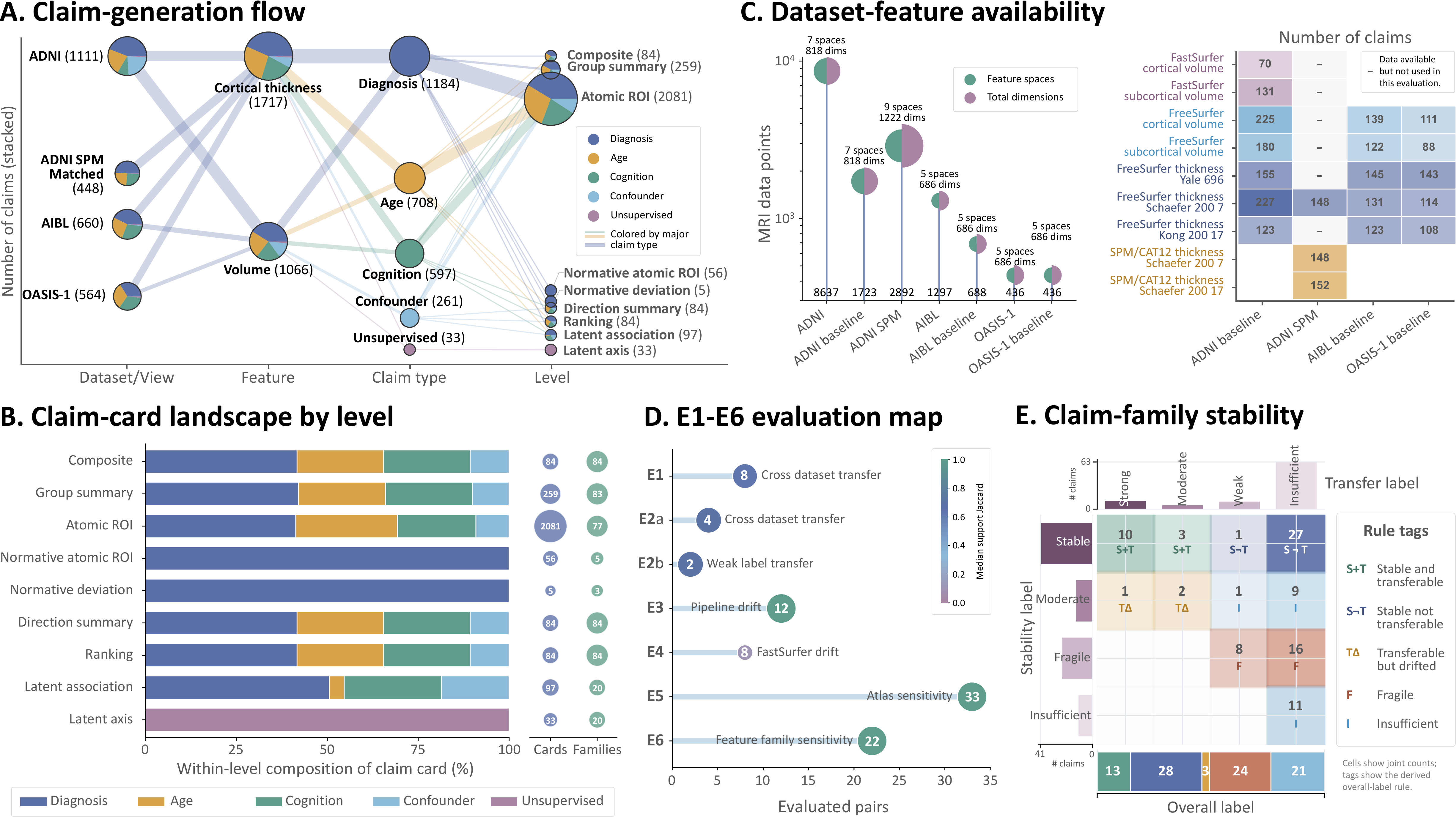}
\caption{Claim landscape and evaluation coverage.
SCS generated 2{,}783 claim cards from 121 claim families and evaluated 89 claim-family pairs across six experiments.
Panels show claim flow, levels, feature availability, experiment coverage, and stability--transfer labels.
Details in Appendix~\ref{app:figure_claim_landscape}.
}
\label{fig:claim_landscape}
\end{figure}

SCS generated 121 claim families and 2{,}783 claim cards from the observed morphometry multiverse.
Figure~\ref{fig:claim_landscape} summarizes the resulting claim landscape across dataset views, feature spaces, claim levels, grouped claim types, and evaluation pairs.
The cards cover diagnosis, age, cognition, confounder associations, and unsupervised structure.
Atomic ROI claims formed the largest layer with 2{,}081 cards, preserving fine-grained regional evidence, while composite, group-summary, ranking, direction-summary, latent-axis, and normative-deviation claims provided higher-level summaries for transfer evaluation and expert review.

Coverage varied by feature family: cortical-thickness features contributed 1{,}717 claim cards because they span multiple atlases and pipeline comparisons, whereas the 1{,}066 volume cards primarily support cross-dataset transfer and FreeSurfer--FastSurfer drift analyses.
The feature-availability map shows how this coverage relates to the measurements available within each dataset view.
Across the landscape, 89 claim-family pairs were evaluated under dataset, pipeline, atlas, and feature-family perturbations.
The resulting map records which scientific claims can be generated and which perturbations can be evaluated, linking the claim inventory to the available comparisons.

\subsection{Stability and transfer define distinct reliability regimes}
\label{sec:results_stability_transfer}

The 89 evaluated claim-family pairs differed in both stability and transfer.
Across all experiments, 41 pairs were stable, 13 moderate, 24 fragile, and 11 insufficient by stability label.
However, stability did not imply transfer: only 13 pairs were stable-and-transferable, whereas 28 were stable-not-transferable and 3 were transferable-but-drifted.
Evidence structure, support-region preservation, and claim-score transfer capture different aspects of reliability.

Table~\ref{tab:results_experiment_summary} summarizes the experiment-level profiles.
ADNI-to-AIBL and ADNI-to-OASIS-1 volume transfer accounted for 11 of the 13 stable-and-transferable pairs, the largest contribution among the experiments.
The FreeSurfer 8 versus SPM/CAT12 thickness comparison was mostly stable at the evidence and support level, but most pairs were stable-not-transferable because target transfer was unavailable, weak, or insufficient.
FreeSurfer 8 versus FastSurfer volume drift had the highest median evidence concordance among the experiments, yet all 8 pairs were fragile, with a median support Jaccard of 0.098.
Atlas and feature-family comparisons were more heterogeneous: coarse support was preserved for some targets, while other comparisons were fragile or insufficient.

\begin{table}[t]
\caption{Experiment-level SCS evaluation summary.
EC denotes median evidence concordance (Spearman); SO denotes median support overlap (Jaccard).}
\label{tab:results_experiment_summary}
\centering
\scriptsize
\begin{tabularx}{\textwidth}{L{0.025\textwidth} L{0.19\textwidth} r L{0.18\textwidth} L{0.13\textwidth} r r Y}
\toprule
\textbf{Exp.} & \textbf{Perturbation} & \textbf{Pairs} & \textbf{Stability} & \textbf{Overall labels}$^*$ & \textbf{EC} & \textbf{SO} & \textbf{Transfer} \\
\midrule
E1
& ADNI$\rightarrow$AIBL volume
& 8
& 6 stable; 2 moderate
& 6 S+T; 1 T$\Delta$; 1 I
& 0.891
& 0.633
& AUROC 0.718; $|\rho|$ 0.352 \\

E2
& ADNI$\rightarrow$OASIS-1 volume$^\dagger$
& 6
& 5 stable; 1 moderate
& 5 S+T; 1 T$\Delta$
& 0.842
& 0.703
& AUROC 0.741; $|\rho|$ 0.634 \\

E3
& FreeSurfer 8 vs. SPM/CAT12 thickness
& 12
& 10 stable; 2 moderate
& 8 S$\neg$T; 2 S+T; 1 T$\Delta$; 1 I
& 0.875
& 1.000
& AUROC 0.704; $|\rho|$ 0.363 \\

E4
& FreeSurfer 8 vs. FastSurfer volume
& 8
& 8 fragile
& 8 F
& 0.903
& 0.098
& AUROC 0.561; $|\rho|$ 0.137 \\

E5
& FreeSurfer 8 atlas sensitivity$^\ddagger$
& 33
& 16 stable; 7 moderate; 10 fragile
& 16 S$\neg$T; 10 F; 7 I
& 0.700
& 1.000
& n/a \\

E6
& FreeSurfer 8 volume vs. thickness$^\ddagger$
& 22
& 4 stable; 1 moderate; 6 fragile; 11 insufficient
& 4 S$\neg$T; 6 F; 12 I
& 0.500
& 1.000
& n/a \\
\bottomrule
\multicolumn{8}{p{0.96\textwidth}}{
\textit{Notes.}
$^*$S+T = stable-and-transferable; S$\neg$T = stable-not-transferable; T$\Delta$ = transferable-but-drifted; F = fragile; I = insufficient.
$^{\dagger}$OASIS-1 transfer uses CDR-derived impairment as a weak-label target.
$^{\ddagger}$SO is computed in grouped anatomical or network space.
}
\end{tabularx}
\end{table}

\begin{figure}[t]
\centering
\includegraphics[width=\textwidth]{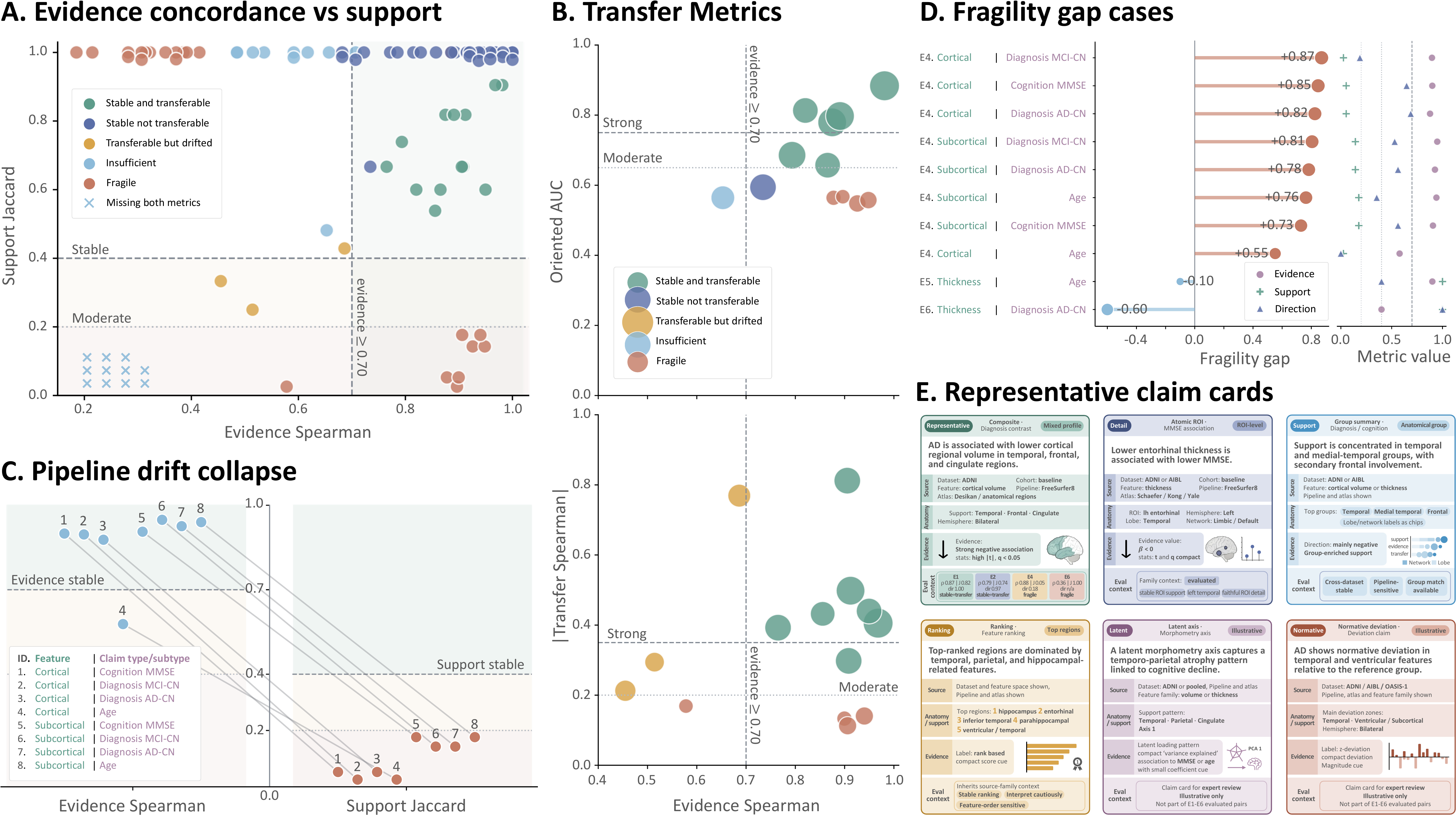}
\caption{Stability-transfer geometry and fragile support.
Evidence concordance and support overlap define distinct claim-reliability regimes.
Under the primary E4 covariate specification, all 8 FreeSurfer 8 versus FastSurfer volume pairs were fragile despite high median evidence Spearman, with substantial changes in supporting regions; Appendix~\ref{app:e4_robustness} reports sensitivity to support rules and covariate adjustment.
Details in Appendix~\ref{app:figure_stability_transfer_fragility}.
}
\label{fig:stability_transfer_fragility}
\end{figure}

\subsection{Pipeline drift exposes high-concordance fragile-support failures}
\label{sec:results_pipeline_drift}

Figure~\ref{fig:stability_transfer_fragility} shows how evidence concordance can coexist with unstable anatomical support.
Evidence concordance and support overlap formed distinct reliability axes: stable-and-transferable claims occupied the high-evidence, high-support regime, whereas fragile claims included cases with preserved global evidence ranking but unstable support regions.
The clearest example was the FreeSurfer 8 versus FastSurfer volume comparison under the primary E4 specification.
All 8 pairs were fragile, with median evidence Spearman of 0.903, support Jaccard of 0.098, and source-support direction retention of 0.546.
This pattern covered cortical and subcortical volume and four targets, although cortical age showed lower evidence concordance (0.578).
An exact hypergeometric lower-tail test of matched-size support overlap, BH-adjusted over the eight comparisons, found below-random overlap for all four cortical pairs ($q=0.012$--$0.040$); the four subcortical pairs were compatible with random overlap ($q=0.448$--$0.563$).

Anatomical-support agreement varied with sample selection and covariate adjustment (Table~\ref{tab:e4_covariate_sensitivity}).
Under the primary covariate specification, restricting the analysis to acquisition-complete observations increased median support Jaccard from 0.098 to 0.194.
Further adjustment for field strength and cohort on these same observations raised it to 0.510, with direction retention reaching 1.000; adding site fixed effects yielded a Jaccard of 0.404, although the CN--AD site models were rank deficient.
E4 support agreement therefore depends on the analysis specification, even when evidence concordance is high.
Additional support-rule, overlap-null, bootstrap, and covariate analyses appear in Appendix~\ref{app:e4_robustness}.

\begin{table}[t]
\caption{E4 sensitivity to sample selection and covariate adjustment (medians across eight FreeSurfer 8--FastSurfer claim-family comparisons).
EC: evidence concordance (Spearman); SO: support overlap (Jaccard); DR: direction retention over source-support regions.}
\label{tab:e4_covariate_sensitivity}
\centering
\small
\begin{tabularx}{\textwidth}{Y r r r}
\toprule
\textbf{Analysis specification} & \hspace{3em}\textbf{EC} & \hspace{3em}\textbf{SO} & \hspace{3em}\textbf{DR} \\
\midrule
Primary covariates, full sample & 0.903 & 0.098 & 0.546 \\
Primary covariates, acquisition-complete subset & 0.885 & 0.194 & 0.656 \\
Additional adjustment for field strength and cohort & 0.892 & 0.510 & 1.000 \\
Additional adjustment for site fixed effects & 0.724 & 0.404 & 1.000 \\
\bottomrule
\end{tabularx}
\vspace{0.2em}
\parbox{\textwidth}{\footnotesize\textit{Note.} Primary covariates are age, sex, education, and intracranial volume, as applicable. The final three rows use identical acquisition-complete observations within each pair, with covariates added cumulatively. CN--AD site models are rank deficient.}
\end{table}

\subsection{Representation sensitivity, anatomy, and review}
\label{sec:results_anatomy_review}

\begin{figure}[t]
\centering
\includegraphics[width=\textwidth]{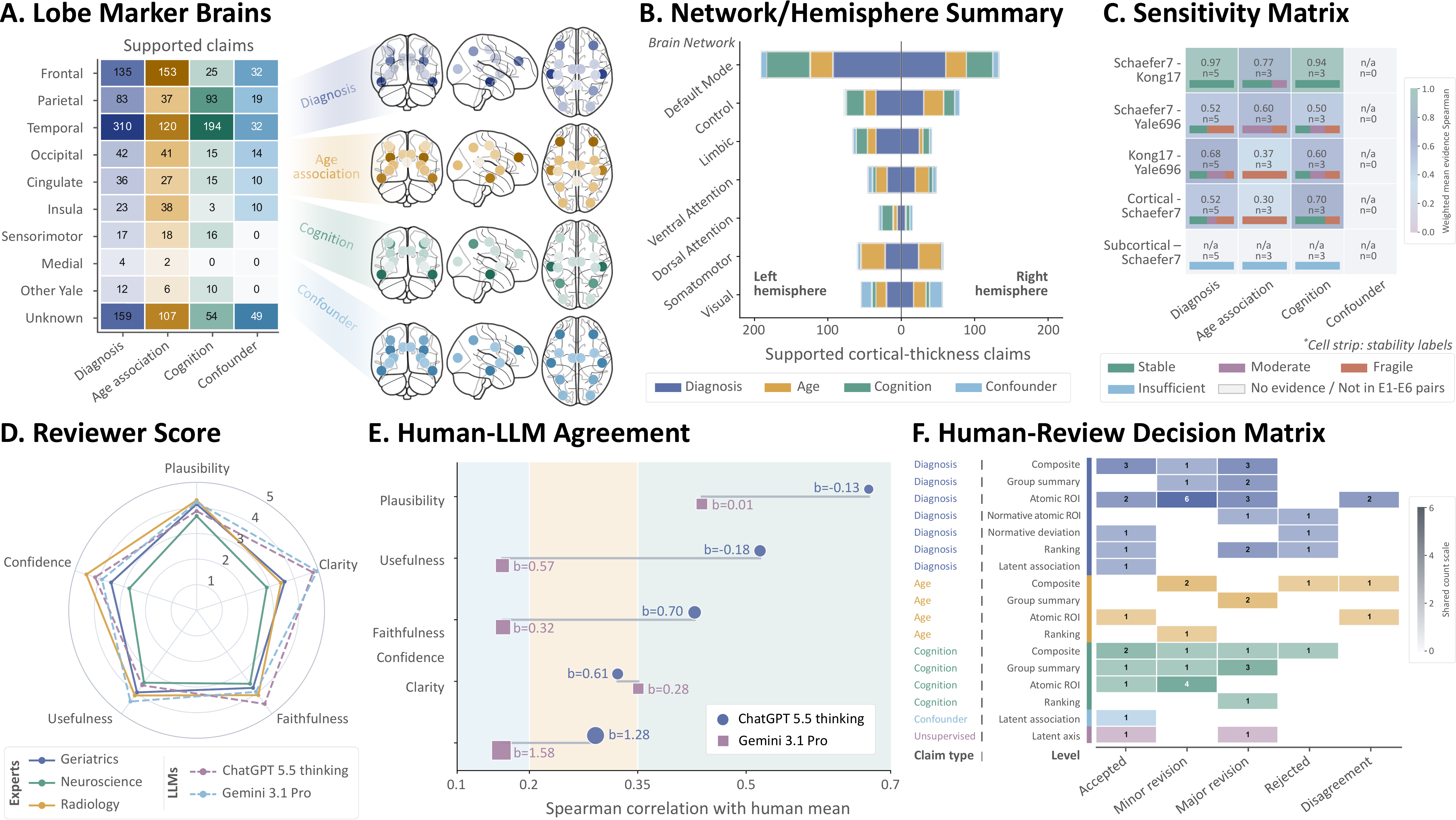}
\caption{Anatomical, representation, and review summaries.
Panels A--C summarize supported claims across anatomical groups and representation changes.
Panels D--F summarize human/LLM review scores, agreement, and decisions.
The anatomical panels summarize the distribution of supporting regions across generated claims.
Details in Appendix~\ref{app:figure_anatomy_review}.
}
\label{fig:anatomy_review}
\end{figure}

Figure~\ref{fig:anatomy_review} summarizes anatomical support, representation sensitivity, and expert/LLM review.
Supported claims frequently involved anatomical groups consistent with established neurodegeneration findings.
Temporal regions were the most prominent lobe-level support group across diagnosis/impairment and cognition targets, and Default-network cortical-thickness claims were the most frequent network-level group.
These patterns are consistent with prior reports of AD-related cortical thinning in temporal, parietal, precuneus, cingulate, and frontal regions, and with default-mode network vulnerability in aging and AD~\citep{dickerson2009cortical,mevel2011default,seoane2024subcorticaldmn}.
The hemisphere summary is exploratory because AD-related asymmetry varies across anatomical regions and studies~\citep{roe2021asymmetric,minkova2017gray}.

The representation-sensitivity panel compares claim stability across atlases and feature families.
Cortical-thickness atlas comparisons retained more stable or moderate structure than volume-versus-thickness comparisons, whereas subcortical-volume-to-thickness rows were frequently insufficient because the feature spaces lacked a meaningful shared cortical ROI mapping.
Diagnosis/impairment, cognition, and age targets also showed different sensitivity patterns.
Anatomical plausibility and stability under atlas or feature-family changes are evaluated separately.

Structured review assessed the scientific usability of selected generated claim cards.
Among the 60 human-reviewed claims, 15 were accepted, 17 required minor revision, 19 required major revision, 5 were rejected, and 4 showed expert disagreement.
Mean plausibility and faithfulness were approximately 4.05 on the 1--5 scale, while mean clarity, usefulness, and confidence were approximately 4.20, 3.95, and 3.85.
Human reviewer agreement was moderate and dimension-dependent, with ICC$(2,k)$ values of 0.690 for plausibility, 0.665 for clarity, and 0.628 for faithfulness, but lower agreement for usefulness and confidence.
Among the LLM judges, ChatGPT had the strongest agreement with human plausibility scores (Spearman correlation 0.670), while both LLM reviewers tended to score clarity higher than the human mean.
Expert comments identified opaque ROI/atlas terminology, insufficient caution, overbroad anatomical summaries, mixed-direction aggregation, and misclassified global or QC features; some rejected claims misidentified features or the direction of an association.
The review identified corrections needed for scientific use of statistically generated candidate claims.

\vspace{-.5em}
\section{Discussion and Limitations}
\label{sec:discussion}
\vspace{-.5em}

SCS evaluates MRI morphometry pipelines through the scientific claims researchers report.
It extends conventional evaluations of measurement agreement, segmentation accuracy, and downstream prediction by examining whether scientific conclusions remain supported, stable, transferable, and usable under realistic perturbations.
Each claim is represented as a typed, machine-evaluable object with a natural-language statement, signed evidence vector, support regions, transfer behavior, perturbation profile, and review status.
Across 10{,}370 MRI scans, SCS generated 2{,}783 claim cards and evaluated 89 claim-family pairs across dataset transfer, weak-label transfer, pipeline drift, atlas sensitivity, and feature-family sensitivity.
These experiments show that claim-level auditing is feasible across the evaluated settings and perturbations without a universal gold standard.

The central empirical finding is the high-concordance fragile-support failure mode under the primary FreeSurfer 8 versus FastSurfer covariate specification.
The four cortical support overlaps were lower than a matched-size random-overlap baseline after BH correction; the four subcortical overlaps were statistically compatible with chance.
On the acquisition-complete subset, median support Jaccard increased from 0.194 to 0.510 after adding field strength and cohort to the covariate model for the same observations.
The full analysis specification, including the dataset, pipeline, covariates, and inclusion criteria, is therefore recorded as the scientific claim's provenance.
SCS records evidence concordance, support overlap, effect direction, and transfer, together with configurable descriptive labels. These measures assess preservation of evidence and anatomical support under a specified perturbation; biological truth and clinical validity require separate evidence.

The observed multiverse comprises 121 analyzed claim-family settings and 89 predefined comparisons, with incomplete coverage of possible combinations.
In neuroimaging studies, datasets differ in the available pipelines, atlases, feature families, targets, and covariate specifications.
SCS evaluates claim stability over the available comparisons and records unavailable, weakly matched, or insufficient comparisons.
This is relevant for studies that reuse processed data, integrate cohorts with different access constraints, compare tools across subsets, or evaluate related but non-identical targets such as diagnosis-derived claims and CDR-derived impairment.

The \texttt{scs\_morph} implementation separates data harmonization, feature construction, claim extraction, stability evaluation, transfer testing, and expert/LLM review into reusable modules.
This modular design allows new datasets, atlases, pipelines, targets, covariate models, and review protocols to be added without changing the core claim-card abstraction.
Human review assesses claim cards for plausibility, clarity, faithfulness, usefulness, and caution.
LLM judges provide scalable auxiliary scores, with agreement assessed against human experts.

The current study has several limitations.
\textbf{First}, this paper focuses on ROI-level morphometry tables; vertex-wise cortical thickness, full surface geometry, voxel-wise maps, curvature, sulcal depth, and mesh-level quality are outside its empirical scope.
\textbf{Second}, the empirical analysis is cross-sectional; longitudinal processing, visit-level dependence, and claim stability of atrophy rates are left for future work.
\textbf{Third}, ADNI and AIBL participant-level source and derived data remain subject to their access and redistribution restrictions; we provide code and aggregate results for independent regeneration under authorized access. OASIS-1 has its own data-use terms and is also omitted from our participant-level artifact release.
\textbf{Fourth}, OASIS-1 uses CDR-derived weak labels, whereas ADNI and AIBL use harmonized CN/MCI/AD labels. OASIS transfer therefore evaluates a related clinical target through weak-label transfer.
\textbf{Fifth}, the expert-review study reports agreement and revision decisions for a targeted set of claims across three clinical and scientific specialties, limiting conclusions about clinical consensus.
Claim-level evaluation tests whether scientific support is preserved under a declared change in dataset, pipeline, representation, target, or covariate specification.

\begin{ack}
\textbf{Funding.}
This work was supported by the Vietnam National Foundation for Science and Technology Development (NAFOSTED) under the project ``Research and Development of a Personalized Machine Learning System for Early Alzheimer's Disease Diagnosis from Adaptive Multimodal Biomarkers'' (Grant No. 102.05-2025.54).

Duy-Thanh Vu and Oliver Y. Ch\'en gratefully acknowledge support from the Swiss National Science Foundation (SNSF; Project No. 3200--0--239967 and Project No. CR00I5--235987).

\textbf{ADNI data acknowledgment.}
Data collection and sharing for the Alzheimer's Disease Neuroimaging Initiative (ADNI) is 
funded by the National Institute on Aging (National Institutes of Health Grant 
U19AG024904). The grantee organization is the Northern California Institute for Research
and Education. In the past, ADNI has also received funding from the National Institute of 
Biomedical Imaging and Bioengineering, the Canadian Institutes of Health Research, and 
private sector contributions through the Foundation for the National Institutes of Health 
(FNIH) including generous contributions from the following: AbbVie, Alzheimer’s Association; 
Alzheimer’s Drug Discovery Foundation; Araclon Biotech; BioClinica, Inc.; Biogen; Bristol-Myers Squibb Company; CereSpir, Inc.; Cogstate; Eisai Inc.; Elan Pharmaceuticals, Inc.; Eli 
Lilly and Company; EuroImmun; F. Hoffmann-La Roche Ltd and its affiliated company
Genentech, Inc.; Fujirebio; GE Healthcare; IXICO Ltd.; Janssen Alzheimer Immunotherapy 
Research \& Development, LLC.; Johnson \& Johnson Pharmaceutical Research \&
Development LLC.; Lumosity; Lundbeck; Merck \& Co., Inc.; Meso Scale Diagnostics, LLC.; 
NeuroRx Research; Neurotrack Technologies; Novartis Pharmaceuticals Corporation; Pfizer 
Inc.; Piramal Imaging; Servier; Takeda Pharmaceutical Company; and Transition
Therapeutics.

\textbf{AIBL data acknowledgment.}
Data used in the preparation of this article were obtained from the Australian Imaging Biomarkers and Lifestyle flagship study of ageing (AIBL).
See \url{https://www.aibl.csiro.au} for further details.
Data were collected by the AIBL study group.
AIBL study methodology has been reported previously \citep{ellis2009australian}.

\textbf{OASIS-1 data acknowledgment.}
Data were provided in part by OASIS-1: Cross-Sectional.
OASIS-1 principal investigators include D. Marcus, R. Buckner, J. Csernansky, and J. Morris.
OASIS-1 was supported by P50 AG05681, P01 AG03991, P01 AG026276, R01 AG021910, P20 MH071616, and U24 RR021382.

\textbf{Competing interests.}
The authors declare no competing interests.

\end{ack}

\bibliographystyle{unsrt}
\bibliography{refs}

\clearpage
\appendix

\counterwithin{table}{section}
\counterwithin{figure}{section}
\renewcommand{\thetable}{\thesection\arabic{table}}
\renewcommand{\thefigure}{\thesection\arabic{figure}}

\section{Code, Data, and Artifact Availability}
\label{app:availability}

\subsection{Code availability}
\label{app:code_availability}

The SCS framework is released as a reproducible code repository:
\begin{center}
\url{https://github.com/catcd/scs-morph}
\end{center}
The repository contains the \texttt{scs\_morph} package, configuration files, command-line scripts, documentation, and aggregate evaluation summaries needed to reproduce the SCS workflow after obtaining authorized access to the required source datasets.
ADNI and AIBL participant-level data and derived participant-level artifacts are excluded from the release, as are OASIS-1 participant-level data and features.
The dataset-specific conditions are described below.
Researchers must obtain the source datasets through the corresponding data-access procedures and regenerate restricted participant-level artifacts locally under the applicable data-use agreements.

\subsection{Data availability}
\label{app:data_availability}

This study uses ADNI, AIBL, and OASIS-1 data.
The ADNI and AIBL source data used in this evaluation were downloaded from the LONI Image and Data Archive (IDA) on January 14, 2026; the respective archives comprise separately updated imaging and metadata files.
The released repository omits participant-level MRI data, clinical variables, scan identifiers, raw feature matrices, derived participant-level morphometry tables, and participant-level claim scores.
Authorized researchers can regenerate the corresponding analyses after obtaining each source dataset under its applicable terms.

ADNI data are available to approved researchers through the LONI Image and Data Archive after application and acceptance of the ADNI Data Use Agreement.\footnote{\url{https://adni.loni.usc.edu/data-samples/adni-data/}}
The ADNI Data Use Agreement prohibits redistribution of individual participant-level raw or derived data and permits publication of summary, non-participant-level results.
It also restricts use of AI tools on ADNI participant-level data unless the tools provide guarantees that the data are safeguarded, contained, and not released to other parties.
Accordingly, this work does not release ADNI participant-level data and does not provide ADNI participant-level data, scan-level features, identifiers, or clinical rows to public-facing AI systems.

AIBL data access is governed by the AIBL Data Use Agreement through the LONI/ADNI access infrastructure.\footnote{\url{https://ida.loni.usc.edu/collaboration/access/appApply.jsp?project=AIBL}}
AIBL data are used only for scientific research purposes, are not redistributed, and are not included as participant-level or derived participant-level records in the released repository.

OASIS-1 data are available under the OASIS data-use terms.\footnote{\url{https://sites.wustl.edu/oasisbrains/home/access/}}
These terms restrict use to noncommercial academic research without prior authorization, prohibit re-identification attempts and facial-recognition or reconstruction uses, and require dataset acknowledgment.
This study releases aggregate analyses and does not include OASIS-1 participant-level MRI data or derived feature matrices in its public artifacts.

\subsection{Released artifacts}
\label{app:released_artifacts}

The released repository includes the framework and configuration files required to reproduce the SCS workflow.
The released artifacts include:

\begin{itemize}
\item configuration files defining datasets, feature spaces, claim extraction, evaluation pairs, thresholds, and expert-review settings;
\item scripts for data audit, harmonization, feature-matrix construction, claim extraction, claim evaluation, and expert-review analysis;
\item documentation describing expected input locations and output artifacts;
\item aggregate result summaries, evaluation-pair summaries, and claim-level summaries that contain no participant identifiers, no scan-level rows, no individual clinical records, and no participant-level feature vectors.
\end{itemize}


\subsection{Reproducibility scope}
\label{app:reproducibility_scope}

Authorized users can reproduce the workflow by obtaining the source datasets through the corresponding access procedures, placing the required raw metadata and morphometry files under the configured raw-data directory, and running the scripts described in Appendix~\ref{app:implementation_details}.
Participant-level artifacts, including clinical tables, and scan-level features, remain subject to the corresponding data-use agreements.
The repository supports reproduction of the SCS methodology and aggregate evaluation outputs under dataset-specific restrictions on participant-level and derived participant-level data.

\section{Framework Details and Claim Objects}
\label{app:framework_details}

\subsection{Analysis settings and observed multiverse}
\label{app:settings}

The main text defines an analysis setting as an element of the product space
$\mathcal{S}=\mathcal{D}\times\mathcal{H}\times\mathcal{P}\times\mathcal{F}\times\mathcal{A}\times\mathcal{T}\times\mathcal{C}$.
The implementation uses the following expanded definition.
An analysis setting is a reproducible specification
\begin{equation}
s =
(d,h,p,f,a,t,c,r,\kappa),
\end{equation}
where $d$ is the dataset view, $h$ is the cohort or subset, $p$ is the processing pipeline, $f$ is the feature family, $a$ is the atlas or region system, $t$ is the target, $c$ is the covariate specification, $r$ is the inclusion rule, and $\kappa$ denotes claim-generation and support criteria.
The main text uses the shorter notation $s=(d,h,p,f,a,t,c)$; the implementation also retains $r$ and $\kappa$ to specify each claim family fully.
Table~\ref{tab:app_analysis_setting_fields} summarizes the analysis-setting fields and their corresponding implementation names.

\begin{table}[!ht]
\caption{Analysis-setting fields used by SCS.
Each setting defines one reproducible morphometry specification from which claim families can be extracted.}
\label{tab:app_analysis_setting_fields}
\centering
\scriptsize
\begin{tabularx}{\textwidth}{L{0.065\textwidth} p{0.045\textwidth} p{0.10\textwidth} X}
\toprule
\textbf{Field} & \textbf{Symbol} & \textbf{Implementation} & \textbf{Meaning} \\
\midrule
dataset or view
& $D$
& \texttt{dataset}
& Dataset or derived view, including ADNI~\citep{petersen2010alzheimers}, AIBL~\citep{ellis2009australian}, OASIS-1~\citep{marcus2007open}, baseline views, and matched views such as the ADNI FreeSurfer--SPM/CAT12 matched subset. \\\hline

cohort or subset
& $H$
& \texttt{cohort}, view tag
& Cohort, data split, or subset identifier, such as ADNI1, ADNI-GO/2, ADNI3, baseline-only views, pooled views, matched subsets, or cohort-specific analysis groups. \\\hline

pipeline
& $P$
& \texttt{pipeline}
& Processing pipeline or software family used to derive morphometric features, including FreeSurfer 8~\citep{fischl2012freesurfer}, Statistical Parametric Mapping~(SPM) with CAT12~\citep{ashburner2000voxel,dias2022cortical}, and FastSurfer~\citep{henschel2020fastsurfer}. \\\hline

feature family
& $F$
& \texttt{feature\_ family}
& Type of morphometric measurement, including subcortical volume, cortical regional volume, and cortical thickness. \\\hline

atlas or region system
& $A$
& \texttt{atlas}
& Region system used to organize ROI features, including Schaefer~\citep{schaefer2018local}, Kong~\citep{kong2022individualized}, Yale Brain Atlas~\citep{collins2025yale}, and FreeSurfer anatomical regions~\citep{desikan2006automated,fischl2004automatically}. \\\hline

target
& $T$
& \texttt{target}
& Scientific target or contrast used to extract claims, including diagnosis, MMSE, age, CDR-derived impairment, demographic variables, or acquisition variables. \\\hline

covariates
& $C$
& \texttt{covariates}
& Adjustment variables included in the ROI-wise association model, such as age, sex, education, intracranial volume, or other setting-specific covariates. \\\hline

inclusion rule
& $R$
& \texttt{inclusion\_ filter}
& Row-level eligibility rule used to define the analysis population before fitting claim-extraction models. \\\hline

criteria
& $\kappa$
& \texttt{criteria}, thresholds
& Claim-generation and support rules, including minimum sample size, top-$k$ support size, effect-strength thresholds, FDR thresholds, and status-label thresholds. \\
\bottomrule
\end{tabularx}
\end{table}

Each setting induces an aligned analysis table
\begin{equation}
\Psi(s)
=
(\mathbf{X}_s,\mathbf{Z}_s,\mathbf{y}_s),
\end{equation}
where $\mathbf{X}_s\in\mathbb{R}^{n_s\times p_s}$ is the morphometry matrix, $\mathbf{Z}_s\in\mathbb{R}^{n_s\times q_s}$ contains metadata and covariates, and $\mathbf{y}_s\in\mathcal{Y}_s^{n_s}$ is the target vector when a supervised target is available.
Rows correspond to scans or scan-level records, and columns of $\mathbf{X}_s$ correspond to anatomical regions or region-level features.
The alignment operator removes rows with missing target, missing covariates, or non-finite feature values when required by the setting.
For unsupervised or normative settings, $\mathbf{y}_s$ may be absent or used only for downstream interpretation.

The observed multiverse is the set of settings instantiated by available data, feature spaces, and experiment configurations:
\begin{equation}
\mathcal{M}_{\mathrm{obs}}
=
\{s_i: i=1,\ldots,m\}.
\end{equation}
This set is a subset of the ideal multiverse $\mathcal{M}$ of all defensible analysis specifications.
In practice, $\mathcal{M}_{\mathrm{obs}}$ is partially observed because not every dataset is processed by every pipeline, not every atlas exists for every feature family, and not every target is available in every cohort.
SCS evaluates claim stability over observed perturbations and marks unsupported comparisons.

\subsection{Feature spaces and setting expansion}
\label{app:feature_spaces}

A feature space specifies how a column in the harmonized data is converted into a matrix $\mathbf{X}_s$.
Formally, a feature space is represented as
\begin{equation}
\omega =
(\mathrm{name}, \mathrm{column}, f, p, a, \mathcal{R}_{\omega}),
\end{equation}
where $\mathrm{name}$ is the feature-space identifier, $\mathrm{column}$ is the source column containing the vectorized features, $f$ is the feature family, $p$ is the pipeline, $a$ is the atlas or region system, and $\mathcal{R}_{\omega}$ is the ordered set of feature names.
Separating the feature matrix from its provenance allows the same extraction procedure to operate on cortical volumes, subcortical volumes, and cortical-thickness atlases.

Setting expansion creates analysis settings by combining dataset views, feature spaces, targets, covariates, and claim-generation rules.
Let $\Omega$ denote the set of feature spaces and $\mathcal{Q}$ the set of claim queries.
A configuration expansion can be written as
\begin{equation}
\mathrm{Expand}(\mathcal{D}_{\mathrm{obs}},\Omega,\mathcal{Q})
\rightarrow
\mathcal{M}_{\mathrm{obs}}.
\end{equation}
Each generated setting must satisfy availability and sample-size constraints.
For a setting $s$, the eligibility indicator is
\begin{equation}
\mathbb{I}_{\mathrm{eligible}}(s)
=
\mathbb{I}
\left[
n_s \geq n_{\min}
\right]
\mathbb{I}
\left[
p_s \geq p_{\min}
\right]
\mathbb{I}
\left[
\mathbf{y}_s \ \mathrm{is\ available\ when\ required}
\right],
\end{equation}
where $n_{\min}$ is the minimum sample size and $p_{\min}$ is the minimum number of usable features.
Settings that fail these constraints are skipped or marked as insufficient.

Evaluation pairs are ordered pairs of settings,
\begin{equation}
\pi = (s_{\mathrm{src}},s_{\mathrm{tgt}},\delta,\mu),
\end{equation}
where $s_{\mathrm{src}}$ is the source setting, $s_{\mathrm{tgt}}$ is the target setting, $\delta$ is the perturbation relation, and $\mu$ is the feature-matching strategy.
The perturbation relation records what changes between settings, such as dataset, weak-label target, pipeline, atlas, feature family, or target.
The matching strategy $\mu$ determines whether evidence vectors are compared through exact ROI identity or through a coarser anatomical grouping.

For exact ROI matching, the shared index set is
\begin{equation}
\mathcal{J}_{st}
=
\mathcal{R}_{s_{\mathrm{src}}}
\cap
\mathcal{R}_{s_{\mathrm{tgt}}},
\end{equation}
where $\mathcal{R}_{s}$ is the feature-name set for setting $s$.
For non-identical feature spaces, SCS uses a grouping map
\begin{equation}
g_s:\mathcal{R}_{s}\rightarrow\mathcal{G},
\end{equation}
where $\mathcal{G}$ may correspond to anatomical groups, lobes, coarse networks, or network-like groups.
The shared comparison space is then
\begin{equation}
\mathcal{G}_{st}
=
g_{s_{\mathrm{src}}}(\mathcal{R}_{s_{\mathrm{src}}})
\cap
g_{s_{\mathrm{tgt}}}(\mathcal{R}_{s_{\mathrm{tgt}}}).
\end{equation}
This permits atlas sensitivity and feature-family sensitivity analyses even when source and target ROI tables do not share identical feature columns.
Exact metric definitions for matched and grouped comparisons are given in Appendix~\ref{app:stability_metrics}.

\subsection{Claim-card schema}
\label{app:claim_schema}

A claim card is the serialized artifact corresponding to the formal claim object in Eq.~\eqref{eq:claim_card}.
The released implementation defines this artifact in the \texttt{ClaimCard} schema and stores claim cards as JSON-serializable records with fields for identity, source setting, anatomy, statistical evidence, support, generated text, and status.
Table~\ref{tab:app_claim_card_fields} groups the claim-card fields by function.

\begin{table}[!ht]
\caption{Claim-card fields.
A claim card stores both the human-readable statement and the machine-readable evidence needed to audit, transfer, and review the claim.}
\label{tab:app_claim_card_fields}
\centering
\scriptsize
\begin{tabularx}{\textwidth}{p{0.1\textwidth} L{0.4\textwidth} Y}
\toprule
\textbf{Field group} & \textbf{Fields} & \textbf{Purpose} \\
\midrule
identity
& \texttt{claim\_id}, \texttt{claim\_family\_id}, \texttt{claim\_level}, \texttt{claim\_type}
& Uniquely identifies the claim and links it to the family of claims extracted from the same analysis setting. \\\hline

source setting
& \texttt{source\_setting}, \texttt{dataset}, \texttt{cohort}, \texttt{n\_features}, \texttt{feature\_space}, \texttt{feature\_family}, \texttt{pipeline}, \texttt{atlas}, \texttt{target}, \texttt{covariates}, \texttt{n\_samples}
& Records the dataset view, feature provenance, target, covariate model, and sample/feature dimensions that produced the claim. \\\hline

anatomy
& \texttt{roi}, \texttt{hemisphere}, \texttt{roi\_group}, \texttt{anatomical\_group}, \texttt{network}, \texttt{lobe}
& Places the claim in an anatomical, network, or group-level coordinate system. \\\hline

evidence
& \texttt{direction}, \texttt{effect\_size}, \texttt{evidence\_metric}, \texttt{evidence\_value}, \texttt{coefficient}, \texttt{standard\_error}, \texttt{t\_value}, \texttt{p\_value}, \texttt{q\_value}
& Stores the signed statistical evidence supporting the claim. \\\hline

support
& \texttt{support\_regions}, \texttt{support\_region\_effects}, \texttt{support\_metrics}, \texttt{criteria\_passed}, \texttt{criteria\_details}
& Records the regions and criteria used to justify the claim. \\\hline

text
& \texttt{claim\_text}, \texttt{short\_claim\_text}, \texttt{machine\_readable\_statement}, \texttt{expert\_review\_hint}
& Links the natural-language claim to a structured statement suitable for evaluation and review. \\\hline

status
& \texttt{status}, \texttt{notes}
& Stores support status and warnings from validation or serialization. \\
\bottomrule
\end{tabularx}
\end{table}

Claim cards are validated before serialization.
The released implementation checks generated text for empty statements, unresolved placeholders, invalid values such as \texttt{nan} or \texttt{None}, and unsafe or overclaiming language.
If a supported claim fails text validation, the validation routine records the warning and can downgrade the claim to weak support.
Text validation checks the statements presented for human review and scientific reporting.

\subsection{Claim levels and claim types}
\label{app:claim_taxonomy}

SCS separates the \emph{level} of a claim from the \emph{type} of scientific target it addresses.
The claim level determines the granularity of the statement, whereas the claim type determines the statistical or scientific question.
A diagnosis contrast, MMSE association, age association, latent axis, or normative-deviation analysis can produce composite, regional, group-level, ranking, or direction-summary claims.
The supported levels, types, and statuses are defined in the \texttt{ClaimCard} schema and used consistently by the claim-family extraction modules.
Table~\ref{tab:app_claim_taxonomy} lists the supported claim levels, claim types, and claim statuses.

\begin{table}[!ht]
\caption{Claim levels, claim types, and claim statuses.
Claim levels specify the granularity of the statement, claim types specify the scientific target, and statuses specify the support or processing state.}
\label{tab:app_claim_taxonomy}
\centering
\scriptsize
\begin{tabularx}{\textwidth}{p{0.075\textwidth} p{0.22\textwidth} Y}
\toprule
\textbf{Category} & \textbf{Value} & \textbf{Meaning} \\
\midrule
claim level
& \texttt{composite}
& A multi-region statement summarizing the main support pattern for one claim family. \\

claim level
& \texttt{atomic\_roi}
& A single-region statement linked to one ROI-level effect. \\

claim level
& \texttt{group\_summary}
& A statement summarizing support concentration in an anatomical group, network, lobe, or related region group. \\

claim level
& \texttt{ranking}
& A statement describing the ranked support regions for a target or contrast. \\

claim level
& \texttt{direction\_summary}
& A statement describing the dominant sign of effects in the support set. \\

claim level
& \texttt{latent\_axis}
& An unsupervised morphometry-axis statement derived without a supervised target. \\

claim level
& \texttt{latent\_axis\_association}
& A statement linking an unsupervised morphometry axis to metadata or labels. \\

claim level
& \texttt{normative\_deviation}
& A statement describing deviation of a target group from a reference group. \\

claim level
& \texttt{normative\_atomic\_roi}
& A single-region normative-deviation statement. \\
\midrule
claim type
& \texttt{diagnosis\_contrast}
& Binary diagnostic contrast such as CN versus AD or CN versus MCI. \\

claim type
& \texttt{cdr\_impairment\_contrast}
& CDR-derived cognitive-impairment contrast, for weak-label settings such as OASIS-1. \\

claim type
& \texttt{mmse\_association}
& Association between morphometry and MMSE. \\

claim type
& \texttt{age\_association}
& Association between morphometry and age. \\

claim type
& \texttt{continuous\_association}
& Generic association with a continuous target. \\

claim type
& \texttt{demographic\_contrast}
& Contrast defined by a demographic variable. \\

claim type
& \texttt{acquisition\_contrast}
& Contrast defined by an acquisition or imaging-protocol variable. \\

claim type
& \texttt{feature\_ranking}
& Ranking of regions by evidence strength. \\

claim type
& \texttt{group\_enrichment}
& Enrichment or concentration of support in an anatomical group. \\

claim type
& \texttt{direction\_summary}
& Summary of the dominant direction of supported effects. \\

claim type
& \texttt{latent\_morphometry\_axis}
& Unsupervised morphometry axis extracted from the feature matrix. \\

claim type
& \texttt{latent\_axis\_association}
& Association between a latent axis and metadata or labels. \\

claim type
& \texttt{normative\_deviation}
& Deviation from a reference distribution or normative group. \\

claim type
& \texttt{normative\_group\_deviation}
& Group-level concentration of normative deviations. \\
\midrule
status
& \texttt{generated}
& Claim card was created but has not yet been assigned a support status. \\

status
& \texttt{supported}
& Claim passes the predefined support criteria. \\

status
& \texttt{weak\_support}
& Claim is generated but should be treated as exploratory or downgraded by validation. \\

status
& \texttt{criteria\_failed}
& Claim does not pass support criteria. \\

status
& \texttt{skipped}
& Claim generation or evaluation was skipped for a documented reason. \\

status
& \texttt{error}
& Claim generation or evaluation encountered an error. \\
\bottomrule
\end{tabularx}
\end{table}

\subsection{Claim templates}
\label{app:claim_templates}

Claim text is generated from structured templates with explicit evidence-to-text mappings.
Templates are populated with the claim type, target labels, direction, feature-family label, region summary, group label, and covariate text.
Each rendered statement is linked to the statistical evidence and source setting.
In the released implementation, the template library is specified in \texttt{configs/claim\_templates.yaml}, while claim-generation settings and support criteria are configured through \texttt{configs/claim\_generation.yaml} and \texttt{configs/claim\_extraction\_all\_main.yaml}.
The templates are rendered by the claim text-generation module and validated before serialization.
Table~\ref{tab:app_claim_templates} gives representative templates used by the framework.

\begin{table}[!ht]
\caption{Representative claim templates.
Templates generate human-readable statements while preserving a direct link to evidence vectors, support regions, and source-setting metadata.}
\label{tab:app_claim_templates}
\centering
\footnotesize
\begin{tabularx}{\textwidth}{p{0.16\textwidth} p{0.22\textwidth} X}
\toprule
\textbf{Template class} & \textbf{Claim type or level} & \textbf{Template pattern} \\
\midrule
diagnosis contrast
& \texttt{diagnosis\_contrast}
& \texttt{\{positive\_label\} is associated with \{direction\} \{feature\_family\_label\} in \{region\_summary\} compared with \{negative\_label\}, after adjustment for \{covariate\_text\}.} \\\hline

MMSE association
& \texttt{mmse\_association}
& \texttt{Higher MMSE is associated with \{direction\} \{feature\_family\_label\} in \{region\_summary\}, after adjustment for \{covariate\_text\}.} \\\hline

age association
& \texttt{age\_association}
& \texttt{Older age is associated with \{direction\} \{feature\_family\_label\} in \{region\_summary\}, after adjustment for \{covariate\_text\}.} \\\hline

ranking
& \texttt{feature\_ranking}
& \texttt{The top-ranked regions for \{target\_label\} are \{top\_regions\}.} \\\hline

group enrichment
& \texttt{group\_enrichment}
& \texttt{The claim signal is concentrated in \{group\_label\}, where \{n\_support\_rois\}/\{n\_group\_rois\} regions meet support criteria.} \\\hline

direction summary
& \texttt{direction\_summary}
& \texttt{Across the support set, \{fraction\_pct\}\% of regions show lower/higher \{feature\_family\_label\} for \{target\_label\}.} \\\hline

latent axis
& \texttt{latent\_morphometry\_ axis}
& \texttt{An unsupervised morphometry axis explains \{explained\_variance\_pct\}\% of variance and is dominated by \{top\_regions\}.} \\\hline

normative deviation
& \texttt{normative\_deviation}
& \texttt{\{target\_group\} shows normative deviation in \{feature\_family\_label\}, most prominently in \{region\_summary\}, relative to the \{reference\_group\} reference.} \\
\bottomrule
\end{tabularx}
\end{table}

Templates constrain the wording of causal and diagnostic statements to the available evidence.
Generated text is rejected or flagged when it contains unresolved placeholders, invalid values, unknown anatomical labels, or banned language.
The weak-support template is used when a generated claim does not meet the predefined evidence threshold and should be treated as exploratory.

\section{Claim Extraction and Evaluation Metrics}
\label{app:technical_details}
\label{app:evaluation_metrics}

\subsection{ROI-wise association models}
\label{app:roiwise_models}

For supervised claim families, SCS fits one covariate-adjusted association model per region to estimate regional target associations~\citep{hastie2009elements}.
Before model fitting, the feature matrix and metadata table are aligned by index or row order, rows with missing target or required covariates are removed, and rows with non-finite feature values are excluded.
Categorical covariates are one-hot encoded with a dropped reference level, and numeric covariates are converted to floating-point values.
Depending on the setting, features and targets can be standardized before model fitting.

For region $j$, the regression model is
\begin{equation}
x_{ij}
=
\alpha_j
+
\beta_j y_i
+
\mathbf{z}_i^\top \boldsymbol{\gamma}_j
+
\epsilon_{ij},
\qquad
i=1,\ldots,n_s,
\qquad
j=1,\ldots,p_s.
\label{eq:app_roiwise_model}
\end{equation}
Here $x_{ij}$ is the morphometry value for scan $i$ and region $j$, $y_i$ is the target variable, and $\mathbf{z}_i$ is the covariate vector.
For binary contrasts, $y_i\in\{0,1\}$ encodes the negative and positive groups.
For continuous associations, $y_i$ is a continuous target such as MMSE or age.
The coefficient $\hat{\beta}_j$ is the region-level target association after covariate adjustment.

Let $\mathbf{d}_i=[1,y_i,\mathbf{z}_i^\top]^\top$ be the design vector and let $\mathbf{D}\in\mathbb{R}^{n_s\times r_s}$ be the design matrix.
For a full-column-rank design $\mathbf{D}$, ordinary least squares estimates
\begin{equation}
\hat{\boldsymbol{\theta}}_j
=
(\mathbf{D}^{\top}\mathbf{D})^{-1}\mathbf{D}^{\top}\mathbf{x}_{j},
\qquad
\hat{\boldsymbol{\theta}}_j
=
(\hat{\alpha}_j,\hat{\beta}_j,\hat{\boldsymbol{\gamma}}_j^\top)^\top.
\label{eq:app_ols}
\end{equation}
The implementation uses the target coefficient and its standard error to compute a $t$-statistic, $p$-value, and signed evidence value for each ROI.
In the primary extraction procedure, regions that cannot be fitted are excluded from support assessment. Separate categorical site-effect sensitivity models use a pseudoinverse estimator for rank-deficient designs; their coefficients are interpreted only at the level of sensitivity summaries (Appendix~\ref{app:e4_robustness}).

Table~\ref{tab:app_claim_extraction_models} summarizes the extraction models used by the implemented claim families.

\begin{table}[!ht]
\caption{Claim extraction models.
Most claim families are derived from ROI-wise association models, while latent-axis and normative-deviation claims use alternative evidence operators.}
\label{tab:app_claim_extraction_models}
\centering
\footnotesize
\begin{tabularx}{\textwidth}{L{0.15\textwidth} L{0.18\textwidth} L{0.18\textwidth} L{0.18\textwidth} Y}
\toprule
\textbf{Claim type} & \textbf{Model} & \textbf{Target} & \textbf{Evidence} & \textbf{Support} \\
\midrule
diagnosis contrast
& ROI-wise OLS with binary target and covariates
& CN vs AD, CN vs MCI, or related binary contrast
& signed coefficient, $t$-statistic, $p$-value, $q$-value
& top-$k$ regions and FDR-supported ROIs \\\hline

CDR impairment contrast
& ROI-wise OLS with binary weak-label target
& CDR-derived impairment
& signed coefficient, $t$-statistic, $p$-value, $q$-value
& top-$k$ regions and FDR-supported ROIs \\\hline

MMSE association
& ROI-wise OLS with continuous target and covariates
& MMSE
& signed coefficient, $t$-statistic, $p$-value, $q$-value
& top-$k$ regions and FDR-supported ROIs \\\hline

age association
& ROI-wise OLS with continuous target and covariates
& age
& signed coefficient, $t$-statistic, $p$-value, $q$-value
& top-$k$ regions and FDR-supported ROIs \\\hline

demographic or acquisition contrast
& ROI-wise OLS with binary or encoded target
& sex, field strength, or acquisition group
& signed coefficient, $t$-statistic, $p$-value, $q$-value
& top-$k$ regions and FDR-supported ROIs \\\hline

latent morphometry axis
& residualization followed by PCA/SVD
& none required
& component loading and explained variance
& top-loading regions \\\hline

latent-axis association
& association between latent score and metadata
& post-hoc metadata target
& correlation or regression coefficient
& metadata-associated latent axis \\\hline

normative deviation
& reference distribution and z-score deviation
& target group versus reference group
& regional deviation score
& top-deviation regions \\
\bottomrule
\end{tabularx}
\end{table}

\subsection{Evidence metrics and FDR adjustment}
\label{app:evidence_fdr}

For each fitted ROI-wise model, SCS stores the estimated target coefficient $\hat{\beta}_j$, its standard error $\widehat{\mathrm{se}}(\hat{\beta}_j)$, a $t$-statistic, and a two-sided $p$-value.
The default signed evidence value is the $t$-statistic,
\begin{equation}
e_j
=
t_j
=
\frac{\hat{\beta}_j}{\widehat{\mathrm{se}}(\hat{\beta}_j)}.
\label{eq:app_t_evidence}
\end{equation}
Other signed evidence definitions, such as standardized coefficients or signed effect sizes, can be used without changing the claim-card structure.
The evidence value $e_j$ must preserve magnitude and direction for consistent evaluation of support, direction, and transfer.

For a family of $p_s$ ROI-wise tests, SCS applies Benjamini--Hochberg false-discovery-rate correction~\citep{benjamini1995controlling}.
Let $p_{(1)}\leq\cdots\leq p_{(p_s)}$ be the ordered $p$-values.
The adjusted value for rank $r$ is computed as
\begin{equation}
q_{(r)}
=
\min_{\ell\geq r}
\left\{
\frac{p_s}{\ell}p_{(\ell)}
\right\},
\label{eq:app_bh}
\end{equation}
with monotonicity enforced over the ordered tests.
The resulting $q_j$ values are stored in each ROI statistic and in atomic claim cards when applicable.
FDR-adjusted evidence is used for support criteria, but top-$k$ support is retained even when no ROI passes a strict FDR threshold, allowing the framework to distinguish weak exploratory claims from unsupported or failed claims.

\subsection{Support selection and criteria}
\label{app:support_criteria}

Given an evidence vector $\mathbf{e}$, SCS first ranks regions by absolute evidence.
The top-$k$ support set is
\begin{equation}
S_k
=
\left\{
j:
|e_j|
\text{ is among the } k \text{ largest values in } \{|e_1|,\ldots,|e_{p_s}|\}
\right\}.
\label{eq:app_topk_support}
\end{equation}
For text generation and transfer evaluation, the implementation stores both the support-region names and their signed effects.
The support set provides interpretability, while the full evidence vector remains available for stability and transfer metrics.

SCS then applies claim-level criteria.
For a generic claim card, the support decision is
\begin{equation}
\mathrm{Pass}(c_s)
=
\mathbb{I}
\left[
n_s\geq n_{\min}
\right]
\mathbb{I}
\left[
|S_k|\geq k_{\min}
\right]
\mathbb{I}
\left[
\overline{|e|}_{S_k}\geq \eta
\right]
\mathbb{I}
\left[
N_{\mathrm{FDR}}\geq m_{\min}
\ \mathrm{or}\
\mathrm{ExploratoryAllowed}
\right],
\label{eq:app_support_pass}
\end{equation}
where $N_{\mathrm{FDR}}$ is the number of ROIs passing the configured FDR threshold.
The exact values of $n_{\min}$, $k_{\min}$, $\eta$, and $m_{\min}$ are configuration-dependent.
Evidence extraction and support assessment are separate steps: a claim can be generated, weakly supported, supported, or rejected by criteria.

Different claim levels use different criteria, summarized in Table~\ref{tab:app_support_criteria}.
Composite claims require sufficient overall support.
Atomic ROI claims require ROI-level statistical support.
Group-summary claims require support concentration within an anatomical or network group.
Ranking claims require a stable ranked support set.
Direction-summary claims require a sufficient fraction of support regions to share the same sign.
Latent-axis and normative-deviation claims use criteria based on variance explained, loading magnitude, reference-group size, or deviation thresholds.

\begin{table}[!ht]
\caption{Support criteria by claim level.
Criteria determine whether a generated statement is marked as supported, weakly supported, or criteria-failed.}
\label{tab:app_support_criteria}
\centering
\footnotesize
\begin{tabularx}{\textwidth}{L{0.18\textwidth} L{0.38\textwidth} Y}
\toprule
\textbf{Claim level} & \textbf{Minimum criteria} & \textbf{Output} \\
\midrule
composite
& Minimum sample size, non-empty top-$k$ support, sufficient mean absolute evidence, and optional FDR support
& Multi-region claim summarizing the main evidence pattern \\\hline

atomic ROI
& ROI-level evidence passes the configured statistical or effect-size threshold
& Single-region claim with coefficient, $t$-statistic, $p$-value, and $q$-value \\\hline

group summary
& A group contains enough support ROIs or sufficient group-level mean absolute evidence
& Anatomical, lobe, or network-level enrichment claim \\\hline

ranking
& Ranked region list can be generated and meets minimum evidence criteria
& Ordered list of regions supporting the target or contrast \\\hline

direction summary
& A configured fraction of support regions shares the same sign
& Directional summary of whether support is predominantly positive, negative, or mixed \\\hline

latent axis
& Component explains sufficient variance and has interpretable high-loading regions
& Unsupervised morphometry-axis claim \\\hline

latent-axis association
& Latent scores show sufficient association with metadata or labels
& Claim linking latent morphometry variation to a target or covariate \\\hline

normative deviation
& Reference group is large enough and target group shows sufficient z-score deviation
& Group-level deviation claim relative to a reference distribution \\\hline

normative atomic ROI
& Individual ROI deviation exceeds configured z-threshold
& Single-region normative-deviation claim \\
\bottomrule
\end{tabularx}
\end{table}

\subsection{Latent-axis and normative-deviation claims}
\label{app:latent_normative}

SCS can extract claims without a supervised target and can characterize deviations from a reference group.
For latent-axis claims, the feature matrix is first optionally residualized with respect to covariates and standardized across regions.
A low-rank decomposition is then applied to the residualized matrix, following the standard principal-component view of dominant multivariate variation~\citep{jolliffe2002principal}.
For the first component,
\begin{equation}
\tilde{\mathbf{X}}_s
\approx
\mathbf{u}_1 \sigma_1 \mathbf{v}_1^\top,
\label{eq:app_svd}
\end{equation}
where $\mathbf{v}_1\in\mathbb{R}^{p_s}$ is the loading vector over regions and $\sigma_1^2/\sum_\ell\sigma_\ell^2$ gives the explained-variance fraction.
The loading vector $\mathbf{v}_1$ is used as the evidence vector $\mathbf{e}$, with support defined by the top-loading regions.
If metadata or labels are available, the component score $\mathbf{u}_1\sigma_1$ can be associated post hoc with age, diagnosis, MMSE, or acquisition variables to generate latent-axis association claims.

Normative-deviation claims compare individuals or groups with a reference distribution, following normative modeling approaches in neuroimaging~\citep{marquand2016understanding}.
SCS defines a reference group $R$ and a target group $G$ within a setting.
For region $j$, the reference mean and standard deviation are
\begin{equation}
\mu_{Rj}
=
\frac{1}{|R|}
\sum_{i\in R}
x_{ij},
\qquad
\sigma_{Rj}
=
\sqrt{
\frac{1}{|R|-1}
\sum_{i\in R}
(x_{ij}-\mu_{Rj})^2
}.
\label{eq:app_reference_stats}
\end{equation}
The regional deviation of the target group is then
\begin{equation}
z_{Gj}
=
\frac{
\bar{x}_{Gj}-\mu_{Rj}
}{
\sigma_{Rj}+\epsilon
},
\qquad
\bar{x}_{Gj}
=
\frac{1}{|G|}
\sum_{i\in G}
x_{ij},
\label{eq:app_normative_z}
\end{equation}
where $\epsilon$ prevents division by zero.
The vector $\mathbf{z}_G=(z_{G1},\ldots,z_{Gp_s})$ becomes the evidence vector for normative-deviation claims.
Support is selected from regions with the largest absolute deviations or from regions exceeding a configured z-threshold.
This procedure generates interpretable claims in weakly labeled settings where a full diagnostic target is unavailable.

\subsection{Exact and group-level feature matching}
\label{app:matching_metrics}

\paragraph{Exact ROI matching.}

For source setting $s$ and target setting $t$, exact ROI matching is used when both settings share a comparable region system.
Let $\mathcal{R}_s$ and $\mathcal{R}_t$ denote the ordered sets of ROI names in the source and target ROI-statistic tables.
The matched ROI set is
\begin{equation}
\mathcal{J}_{st}
=
\mathcal{R}_s
\cap
\mathcal{R}_t.
\label{eq:app_exact_roi_set}
\end{equation}
All exact-match stability metrics are computed only on $\mathcal{J}_{st}$.
If $|\mathcal{J}_{st}|<5$, the comparison is marked as insufficient because evidence concordance and support overlap are not meaningful with too few matched features.
Exact matching is used for comparisons such as matched FreeSurfer--SPM Schaefer features or FreeSurfer--FastSurfer features with shared anatomical names.

For a matched ROI $j\in\mathcal{J}_{st}$, the source and target statistics are denoted $e_{s,j}$ and $e_{t,j}$.
The default signed evidence is the ROI-wise $t$-statistic, but the same definitions apply to coefficients or other signed evidence values.
Support sets are derived from the source and target ROI-statistic tables using the same top-$k$ rule unless otherwise specified.

\paragraph{Group-level matching for non-identical feature spaces.}

Some comparisons do not admit exact ROI matching.
For example, Schaefer, Kong, and Yale cortical-thickness atlases have different parcels, and regional volume and cortical thickness use different region systems.
SCS maps ROI names to a common anatomical or network-level space for group-level matching.
Let
\begin{equation}
g_s:\mathcal{R}_s\rightarrow\mathcal{G},
\qquad
g_t:\mathcal{R}_t\rightarrow\mathcal{G}
\label{eq:app_group_maps}
\end{equation}
map source and target ROIs to shared groups such as \texttt{network\_7\_like}, \texttt{lobe}, \texttt{roi\_group}, or \texttt{anatomical\_group}.
The shared group set is
\begin{equation}
\mathcal{G}_{st}
=
g_s(\mathcal{R}_s)
\cap
g_t(\mathcal{R}_t).
\label{eq:app_shared_group_set}
\end{equation}

For each group $g\in\mathcal{G}_{st}$, SCS aggregates signed evidence by averaging over ROIs assigned to the group:
\begin{equation}
\bar{e}_{s,g}
=
\frac{1}{|\mathcal{R}_{s,g}|}
\sum_{j\in\mathcal{R}_{s,g}}
e_{s,j},
\qquad
\mathcal{R}_{s,g}
=
\{j\in\mathcal{R}_s:g_s(j)=g\}.
\label{eq:app_group_evidence}
\end{equation}
The same operation is applied to the target setting.
Group-level support sets are then defined by ranking groups according to $|\bar{e}_{s,g}|$ or group-level mean absolute evidence.
The resulting group-level evidence permits comparison across non-identical feature spaces.

\subsection{Evidence, support, direction, and magnitude stability}
\label{app:stability_metrics}

SCS computes stability as a collection of complementary metrics.
Table~\ref{tab:app_stability_metrics} summarizes the metrics used for exact ROI and group-level comparisons.
Group-level comparisons use the same metrics, replacing ROI evidence $e_j$ with group evidence $\bar{e}_g$.

\begin{table}[!ht]
\caption{Stability metrics used by SCS.
The metrics separate evidence agreement, support-region preservation, direction consistency, and effect-size drift.}
\label{tab:app_stability_metrics}
\centering
\footnotesize
\begin{tabularx}{\textwidth}{L{0.20\textwidth} L{0.42\textwidth} Y}
\toprule
\textbf{Metric} & \textbf{Formula} & \textbf{Interpretation} \\
\midrule
evidence Spearman
& $\rho_{\mathrm{Spearman}}(\mathbf{e}_s[\mathcal{J}_{st}],\mathbf{e}_t[\mathcal{J}_{st}])$
& Rank concordance of signed evidence vectors over matched ROIs. \\\hline

evidence Pearson
& $\rho_{\mathrm{Pearson}}(\mathbf{e}_s[\mathcal{J}_{st}],\mathbf{e}_t[\mathcal{J}_{st}])$
& Linear concordance of signed evidence vectors over matched ROIs. \\\hline

weighted Jaccard of absolute evidence
& $\dfrac{\sum_{j\in\mathcal{J}_{st}}\min(|e_{s,j}|,|e_{t,j}|)}{\sum_{j\in\mathcal{J}_{st}}\max(|e_{s,j}|,|e_{t,j}|)}$
& Overlap of evidence magnitude profiles, giving larger weight to high-evidence ROIs. \\\hline

top-$k$ rank overlap
& $\dfrac{|R_k(s)\cap R_k(t)|}{k}$
& Fraction of top-ranked evidence regions shared by source and target. \\\hline

support Jaccard
& $\dfrac{|S_k(c_s)\cap S_k(c_t)|}{|S_k(c_s)\cup S_k(c_t)|}$
& Overlap of the source and target support regions. \\\hline

source support overlap
& $\dfrac{|S_k(c_s)\cap S_k(c_t)|}{|S_k(c_s)|}$
& Fraction of source-support regions retained in the target support set. \\\hline

target support overlap
& $\dfrac{|S_k(c_s)\cap S_k(c_t)|}{|S_k(c_t)|}$
& Fraction of target-support regions already present in the source support set. \\\hline

direction retention
& $\dfrac{1}{|S_k(c_s)|}\sum_{j\in S_k(c_s)}\mathbb{I}[\operatorname{sign}(e_{s,j})=\operatorname{sign}(e_{t,j})]$
& Fraction of source-support regions with preserved effect direction. \\\hline

sign-flip fraction
& $1-DR_k(s,t)$
& Fraction of source-support regions whose effect direction changes sign. \\\hline

magnitude drift over matched features
& $\dfrac{1}{|\mathcal{J}_{st}|}\sum_{j\in\mathcal{J}_{st}} |z(e_{s,j})-z(e_{t,j})|$
& Mean standardized evidence change over all matched ROIs. \\\hline

magnitude drift over source support
& $\dfrac{1}{|S_k(c_s)|}\sum_{j\in S_k(c_s)} |z(e_{s,j})-z(e_{t,j})|$
& Mean standardized evidence change over the regions used to justify the source claim. \\\hline

group evidence concordance
& $\rho(\bar{\mathbf{e}}_s[\mathcal{G}_{st}],\bar{\mathbf{e}}_t[\mathcal{G}_{st}])$
& Evidence concordance after aggregation to a shared group space. \\\hline

group support Jaccard
& $\dfrac{|S_k^{\mathcal{G}}(c_s)\cap S_k^{\mathcal{G}}(c_t)|}{|S_k^{\mathcal{G}}(c_s)\cup S_k^{\mathcal{G}}(c_t)|}$
& Overlap of high-evidence anatomical, network, or lobe groups. \\\hline

group direction retention
& $\dfrac{1}{|S_k^{\mathcal{G}}(c_s)|}\sum_{g\in S_k^{\mathcal{G}}(c_s)}\mathbb{I}[\operatorname{sign}(\bar{e}_{s,g})=\operatorname{sign}(\bar{e}_{t,g})]$
& Direction preservation over source-support groups. \\
\bottomrule
\end{tabularx}
\end{table}

Here $R_k(s)$ denotes the top-$k$ ranked ROI set by absolute evidence, and $S_k(c_s)$ denotes the support set attached to the source claim card.
The notation $z(\cdot)$ denotes standardization within the compared evidence vector before computing magnitude drift.
For group-level metrics, $S_k^{\mathcal{G}}(c_s)$ denotes the top-$k$ support groups.

\subsection{Claim-score transfer}
\label{app:transfer_metrics}

Claim-score transfer evaluates whether a source claim remains useful in target data without refitting.
The source claim defines a weight vector $\mathbf{w}_s$ over matched target features.
By default, the weight vector is derived from source coefficients or signed evidence values and normalized:
\begin{equation}
\mathbf{w}_s
=
\frac{\tilde{\mathbf{e}}_s}{\|\tilde{\mathbf{e}}_s\|_2+\epsilon},
\label{eq:app_transfer_weight}
\end{equation}
where $\tilde{\mathbf{e}}_s$ is the source evidence restricted to matched features.
For target feature matrix $\mathbf{X}_t$, the transferred claim score is
\begin{equation}
\mathbf{u}^{(t)}
=
\mathbf{X}_t\mathbf{w}_s.
\label{eq:app_transfer_score}
\end{equation}
The target feature matrix is standardized before scoring when configured.

Table~\ref{tab:app_transfer_metrics} summarizes the transfer metrics.
For binary targets, SCS orients the AUROC so that values above $0.5$ indicate alignment with the configured positive class.
Cohen's $d$ is computed as the standardized mean difference between claim scores in the positive and negative target groups.
For continuous targets, SCS reports both Spearman and Pearson correlations between the transferred claim score and the target.

\begin{table}[!ht]
\caption{Transfer metrics used by SCS.
Transfer evaluates whether a source claim remains useful in target data without re-estimating the claim.}
\label{tab:app_transfer_metrics}
\centering
\footnotesize
\begin{tabularx}{\textwidth}{L{0.15\textwidth} L{0.16\textwidth} Y}
\toprule
\textbf{Target type} & \textbf{Metric} & \textbf{Interpretation} \\
\midrule
binary
& oriented AUROC
& Discrimination of the target contrast by the transferred claim score, oriented so higher values favor the configured positive class. \\\hline

binary
& Cohen's $d$
& Standardized difference in transferred claim score between positive and negative target groups. \\\hline

continuous
& Spearman correlation
& Rank association between transferred claim score and continuous target. \\\hline

continuous
& Pearson correlation
& Linear association between transferred claim score and continuous target. \\\hline

none or unsupported
& not applicable
& Transfer is not computed when no target is defined or feature matching is not available. \\
\bottomrule
\end{tabularx}
\end{table}

\subsection{Stability, transfer, and overall labels}
\label{app:labels}

SCS assigns descriptive labels after computing the metric profile.
The labels summarize the underlying metrics for interpretation.
Stability and transfer labels use prespecified, configurable thresholds, with E4 evaluated under these same thresholds.
The thresholds used in the experiments are summarized in Table~\ref{tab:app_label_rules}, and E4 label sensitivity is reported in Appendix~\ref{app:e4_robustness}.

\begin{table}[!ht]
\caption{Label assignment rules used in the experiments.
Labels accompany the underlying stability and transfer metrics and summarize the multidimensional profile.}
\label{tab:app_label_rules}
\centering
\scriptsize
\begin{tabularx}{\textwidth}{L{0.2\textwidth} L{0.35\textwidth} Y}
\toprule
\textbf{Label} & \textbf{Condition} & \textbf{Meaning} \\
\midrule
insufficient stability
& fewer than 5 matched ROIs or fewer than 5 matched groups
& Too few matched units to evaluate stability reliably. \\\hline

stable
& evidence Spearman $\geq 0.70$, support Jaccard $\geq 0.40$, and direction retention $\geq 0.80$
& Evidence ranking, support set, and direction are all strongly preserved. \\\hline

moderate
& evidence Spearman $\geq 0.40$, support Jaccard $\geq 0.20$, and direction retention $\geq 0.60$
& The claim is partially preserved but shows nontrivial drift. \\\hline

fragile
& evaluated but does not meet moderate stability criteria
& The claim changes substantially under the perturbation. \\\hline

strong binary transfer
& oriented AUROC $\geq 0.75$ and $|d|\geq 0.50$
& The transferred claim score strongly separates the target binary groups. \\\hline

moderate binary transfer
& oriented AUROC $\geq 0.65$ and $|d|\geq 0.30$
& The transferred claim score moderately separates the target binary groups. \\\hline

weak binary transfer
& binary transfer evaluated but below moderate thresholds
& The claim score has limited binary transfer utility. \\\hline

strong continuous transfer
& $|\rho_{\mathrm{Spearman}}|\geq 0.35$
& The transferred claim score has strong rank association with the target. \\\hline

moderate continuous transfer
& $|\rho_{\mathrm{Spearman}}|\geq 0.20$
& The transferred claim score has moderate rank association with the target. \\\hline

weak continuous transfer
& continuous transfer evaluated but below moderate threshold
& The claim score has limited continuous transfer utility. \\\hline

not applicable transfer
& transfer target is absent or transfer is not defined
& Stability can still be evaluated, but transfer is not part of the profile. \\\hline

stable and transferable
& stability label is stable and transfer label is strong or moderate
& The claim is preserved and useful in the target setting. \\\hline

stable not transferable
& stability label is stable but transfer is weak, not applicable, or insufficient
& The claim is structurally preserved but does not show useful transfer. \\\hline

transferable but drifted
& stability label is moderate or fragile and transfer label is strong or moderate
& The claim score transfers, but the supporting evidence pattern changes. \\\hline

overall fragile
& stability label is fragile and transfer is not strong or moderate
& The claim is not reliably preserved under the perturbation. \\\hline

overall insufficient
& stability label is insufficient, or stability label is moderate and transfer is weak, not applicable, or insufficient
& The pair lacks enough matched information or sufficient joint stability--transfer evidence for a stronger overall conclusion. \\
\bottomrule
\end{tabularx}
\end{table}

Stability and transfer can diverge.
A claim can preserve its support regions without producing a useful target score, or it can transfer as a score while drifting anatomically.
SCS reports the full metric profile alongside the descriptive labels to retain these distinctions.

\section{Empirical Design and Reproducibility Details}
\label{app:empirical_design_details}
\label{app:datasets_features}

\subsection{Dataset views and targets}
\label{app:dataset_views_targets}

Table~\ref{tab:app_dataset_views} summarizes the dataset views used in the empirical evaluation.
The total analysis set contains 10{,}370 MRI scans across ADNI, AIBL, and OASIS-1.
Per-view scan counts are computed from the harmonized metadata and feature-matrix inventories.
OASIS-1 is treated separately from ADNI and AIBL because it is not used with the same CN/MCI/AD harmonization; when a weak-label contrast is needed, OASIS-1 uses CDR-derived cognitive impairment.

\begin{table}[!ht]
\caption{Dataset views used by SCS.
The table separates full, baseline, and matched views because different evaluation experiments require different levels of alignment.}
\label{tab:app_dataset_views}
\centering
\scriptsize
\begin{tabularx}{\textwidth}{L{0.1\textwidth} L{0.08\textwidth} L{0.09\textwidth} L{0.2\textwidth} Y}
\toprule
\textbf{Dataset view} & \textbf{Dataset} & \textbf{Scan count} & \textbf{Targets} & \textbf{Role in evaluation} \\
\midrule
\texttt{ADNI}
& ADNI~\citep{petersen2010alzheimers}
& 8{,}637
& CN/MCI/AD, MMSE, age, sex, field strength, cohort
& Full ADNI view used for feature construction and claim-generation inventory. \\

\texttt{ADNI\_ baseline}
& ADNI~\citep{petersen2010alzheimers}
& 1{,}723
& CN/MCI/AD, MMSE, age, sex, field strength, cohort
& Baseline ADNI view used as the main source for volume transfer, FastSurfer drift, atlas sensitivity, and feature-family sensitivity. \\

\texttt{AIBL}
& AIBL~\citep{ellis2009australian}
& 1{,}297
& CN/MCI/AD-compatible flags, MMSE, age, sex, imaging protocol
& Full AIBL view used for feature construction and claim-generation inventory. \\

\texttt{AIBL\_ baseline}
& AIBL~\citep{ellis2009australian}
& 688
& CN/MCI/AD-compatible flags, MMSE, age, sex, imaging protocol
& External target for ADNI-to-AIBL volume transfer and atlas/feature-family sensitivity. \\

\texttt{OASIS-1}
& OASIS-1~\citep{marcus2007open}
& 436
& CDR-derived impairment, MMSE, age, sex, education
& Full OASIS-1 view used for feature construction and weak-label claim-generation inventory. \\

\texttt{OASIS-1\_ baseline}
& OASIS-1~\citep{marcus2007open}
& 436
& CDR-derived impairment, MMSE, age, sex, education
& External weak-label target for ADNI-to-OASIS transfer and atlas/feature-family sensitivity. \\

\texttt{ADNI\_SPM\_ MATCHED}
& ADNI~\citep{petersen2010alzheimers}
& 2{,}892
& CN/MCI/AD, MMSE, age
& Matched FreeSurfer--SPM/CAT12 cortical-thickness view used for direct pipeline-drift evaluation. \\
\bottomrule
\end{tabularx}
\end{table}

\subsection{Feature spaces and matched matrix sizes}
\label{app:feature_inventory}

\paragraph{Feature spaces.}
Table~\ref{tab:app_feature_spaces} lists the feature spaces used by the framework.
Each feature space is defined by a feature family, processing pipeline, atlas or region system, and the dataset views in which it is available.
Feature vectors are parsed into scan-by-region matrices and paired with harmonized metadata before claim extraction.

\begin{table}[!ht]
\caption{Feature spaces used by SCS.
The inventory distinguishes volume, cortical-thickness, pipeline-perturbation, and matched feature spaces.}
\label{tab:app_feature_spaces}
\centering
\scriptsize
\begin{tabularx}{\textwidth}{L{0.17\textwidth} L{0.1\textwidth} L{0.15\textwidth} L{0.18\textwidth} Y}
\toprule
\textbf{Feature space} & \textbf{Family} & \textbf{Pipeline} & \textbf{Atlas or region system} & \textbf{Available dataset views} \\
\midrule
\texttt{fs8\_cortical\_ volume}
& cortical regional volume
& FreeSurfer 8~\citep{fischl2012freesurfer}
& FreeSurfer cortical anatomical regions~\citep{desikan2006automated,fischl2004automatically}
& ADNI, ADNI baseline, AIBL, AIBL baseline, OASIS-1, OASIS-1 baseline \\\hline

\texttt{fs8\_subcortical\_ volume}
& subcortical volume
& FreeSurfer 8~\citep{fischl2012freesurfer}
& FreeSurfer subcortical anatomical regions
& ADNI, ADNI baseline, AIBL, AIBL baseline, OASIS-1, OASIS-1 baseline \\\hline

\texttt{fs8\_thickness\_ schaefer200\_7}
& cortical thickness
& FreeSurfer 8~\citep{fischl2012freesurfer}
& Schaefer 200 / 7-network atlas~\citep{schaefer2018local}
& ADNI, ADNI baseline, AIBL, AIBL baseline, OASIS-1, OASIS-1 baseline, ADNI SPM-matched \\\hline

\texttt{fs8\_thickness\_ kong200\_17}
& cortical thickness
& FreeSurfer 8~\citep{fischl2012freesurfer}
& Kong 200 / 17-network atlas~\citep{kong2022individualized}
& ADNI, ADNI baseline, AIBL, AIBL baseline, OASIS-1, OASIS-1 baseline \\\hline

\texttt{fs8\_thickness\_ yale696}
& cortical thickness
& FreeSurfer 8~\citep{fischl2012freesurfer}
& Yale Brain Atlas~\citep{collins2025yale}
& ADNI, ADNI baseline, AIBL, AIBL baseline, OASIS-1, OASIS-1 baseline \\\hline

\texttt{spm\_thickness\_ schaefer200\_7}
& cortical thickness
& SPM/CAT12~\citep{ashburner2000voxel,dias2022cortical}
& Schaefer 200 / 7-network atlas~\citep{schaefer2018local}
& ADNI SPM-matched \\\hline

\texttt{spm\_thickness\_ schaefer200\_17}
& cortical thickness
& SPM/CAT12~\citep{ashburner2000voxel,dias2022cortical}
& Schaefer 200 / 17-network mapping~\citep{schaefer2018local}
& ADNI SPM-matched \\\hline

\texttt{fastsurfer\_ cortical\_volume}
& cortical regional volume
& FastSurfer~\citep{henschel2020fastsurfer}
& FreeSurfer-compatible cortical anatomical regions
& ADNI baseline \\\hline

\texttt{fastsurfer\_ subcortical\_ volume}
& subcortical volume
& FastSurfer~\citep{henschel2020fastsurfer}
& FreeSurfer-compatible subcortical anatomical regions
& ADNI baseline \\
\bottomrule
\end{tabularx}
\end{table}

\paragraph{Matched matrix sizes.}
\label{app:matched_matrix_sizes}

The ADNI SPM-matched view is used for direct cortical-thickness pipeline-drift evaluation.
It contains the same 2{,}892 scans with aligned FreeSurfer 8 and SPM/CAT12 Schaefer thickness representations.
Table~\ref{tab:app_matched_matrix_sizes} gives the matched matrix sizes.

\begin{table}[!ht]
\caption{Matched matrix sizes for the ADNI FreeSurfer--SPM/CAT12 cortical-thickness comparison.
All three matrices contain the same 2{,}892 matched scans and 202 columns.}
\label{tab:app_matched_matrix_sizes}
\centering
\begin{tabularx}{\textwidth}{L{0.35\textwidth} L{0.09\textwidth} L{0.09\textwidth} Y}
\toprule
\textbf{Dataset view / feature space} & \textbf{Rows} & \textbf{Columns} & \textbf{Use} \\
\midrule
\texttt{ADNI\_SPM\_MATCHED/ spm\_thickness\_schaefer200\_7}
& 2{,}892
& 202
& SPM/CAT12 Schaefer 7-network thickness target for pipeline-drift evaluation. \\

\texttt{ADNI\_SPM\_MATCHED/ spm\_thickness\_schaefer200\_17}
& 2{,}892
& 202
& SPM/CAT12 Schaefer 17-network thickness target for network-resolution drift evaluation. \\

\texttt{ADNI\_SPM\_MATCHED/ fs8\_thickness\_schaefer200\_7}
& 2{,}892
& 202
& FreeSurfer 8 Schaefer 7-network thickness source for matched pipeline comparison. \\
\bottomrule
\end{tabularx}
\end{table}

\subsection{Claim Generation Inventory}
\label{app:claim_generation_inventory}

\subsubsection{Claim-generation settings}
\label{app:claim_generation_settings}

Claim generation is performed by expanding configured analysis settings across dataset views, feature spaces, targets, covariates, claim types, and support criteria.
Each expanded setting defines one claim family.
The implementation then applies the appropriate claim extractor to produce one or more claim cards from that family.
The resulting inventory contains 121 claim families and 2{,}783 claim cards.

The claim-generation configuration has three components.
First, dataset and feature-space configurations define which morphometry matrices are available.
Second, claim-extraction configurations define targets, covariates, inclusion filters, claim types, and minimum sample requirements.
Third, claim-generation settings define which claim levels are generated, including composite, atomic ROI, group-summary, ranking, direction-summary, latent-axis, and normative-deviation claims.

New datasets, feature spaces, targets, and claim templates can be added through configuration without changing the core extraction code.

\subsubsection{Claim-family and claim-card inventory}
\label{app:claim_family_card_inventory}

Table~\ref{tab:app_claim_generation_summary} summarizes the claim-generation output used in the paper.
The full claim-card table and claim-family table are released as structured output artifacts.

\begin{table}[!ht]
\caption{Claim-generation summary.
Each analysis setting produces one claim family, and each claim family can produce multiple claim cards at different levels of granularity.}
\label{tab:app_claim_generation_summary}
\centering
\begin{tabularx}{\textwidth}{L{0.26\textwidth} Y}
\toprule
\textbf{Quantity} & \textbf{Value} \\
\midrule
Analysis settings
& 121 \\

Claim families
& 121 \\

Claim cards
& 2{,}783 \\

Claim-card output
& \texttt{data/processed/claim\_cards/claim\_cards.jsonl} and \texttt{claim\_cards\_summary.csv} \\

Claim-family output
& \texttt{data/processed/claim\_cards/claim\_families.jsonl} and \texttt{claim\_family\_summary.csv} \\

Per-setting ROI statistics
& \texttt{data/processed/claim\_cards/per\_setting/} \\
\bottomrule
\end{tabularx}
\end{table}

\clearpage
\subsubsection{Generated claim levels and claim types}
\label{app:generated_claim_levels_types}

The generated claim inventory covers supervised, weak-label, unsupervised, and normative settings.
Table~\ref{tab:app_generated_claim_levels_types} lists the claim levels and claim types included in the generated pool.
Counts by dataset, feature space, target, claim level, and claim type are reported in the released claim-card summary table.

\begin{table}[!ht]
\caption{Generated claim levels and claim types.
The claim-generation inventory includes multiple granularities of scientific statements from the same underlying evidence models.}
\label{tab:app_generated_claim_levels_types}
\centering
\begin{tabularx}{\textwidth}{L{0.1\textwidth} Y}
\toprule
\textbf{Category} & \textbf{Generated values} \\
\midrule
claim levels
& \texttt{composite}, \texttt{atomic\_roi}, \texttt{group\_summary}, \texttt{ranking}, \texttt{direction\_summary}, \texttt{latent\_axis}, \texttt{latent\_axis\_association}, \texttt{normative\_deviation}, \texttt{normative\_atomic\_roi} \\

claim types
& \texttt{diagnosis\_contrast}, \texttt{cdr\_impairment\_contrast}, \texttt{mmse\_association}, \texttt{age\_association}, \texttt{continuous\_association}, \texttt{demographic\_contrast}, \texttt{acquisition\_contrast}, \texttt{feature\_ranking}, \texttt{group\_enrichment}, \texttt{direction\_summary}, \texttt{latent\_morphometry\_axis}, \texttt{latent\_axis\_association}, \texttt{normative\_deviation}, \texttt{normative\_group\_deviation} \\
\bottomrule
\end{tabularx}
\end{table}

\subsubsection{Claim-generation outputs used downstream}
\label{app:claim_generation_downstream_outputs}

Claim-generation outputs are used in review and evaluation.
First, claim cards provide the candidate pool for expert and LLM review.
Second, claim families provide the source and target objects used in stability and transfer evaluation.
Third, per-setting ROI statistics provide the evidence vectors, support regions, and anatomical annotations used to compute exact or group-level stability metrics.

SCS evaluates claims at the family level and retains the individual claim cards.
For example, a diagnosis-contrast family can contain a composite claim, atomic ROI claims, group-summary claims, a feature-ranking claim, and a direction-summary claim.
The family-level evidence is then compared across settings, while selected claim cards are exported for human and LLM review.

\subsection{Evaluation Experiment Details}
\label{app:evaluation_experiment_details}

\subsubsection{Evaluation-pair construction}
\label{app:evaluation_pair_construction}

SCS evaluates stability and transfer through ordered source--target pairs.
Each evaluation pair is defined as
\begin{equation}
\pi =
(s_{\mathrm{src}},s_{\mathrm{tgt}},\delta,\mu,\tau_{\mathrm{src}},\tau_{\mathrm{tgt}}),
\end{equation}
where $s_{\mathrm{src}}$ is the source setting, $s_{\mathrm{tgt}}$ is the target setting, $\delta$ is the perturbation type, $\mu$ is the ROI matching strategy, and $\tau_{\mathrm{src}}$ and $\tau_{\mathrm{tgt}}$ are the source and target claim types.
The evaluation-pair configuration specifies the source and target dataset views, feature spaces, claim types, targets, matching strategy, and transfer target when applicable.
The expanded evaluation table contains one row per source--target claim-family comparison.

Evaluation requires both source and target claim families to be available.
For each pair, the evaluator loads the corresponding ROI-level statistics, claim-family metadata, support regions, and feature matrices when transfer scoring is required.
Pairs are evaluated using exact ROI matching when source and target feature spaces share comparable ROI names.
Pairs are evaluated using group-level matching when source and target feature spaces do not share direct ROI correspondence.
Pairs with no valid matched features, missing claim families, or missing ROI statistics are marked as skipped or insufficient.

In the reported experiments, 89 evaluation pairs were expanded and all 89 were evaluated.
No configured pair was skipped and no pair produced a processing error.

\clearpage
\subsubsection{Implemented evaluation experiments}
\label{app:implemented_evaluation_experiments}

Table~\ref{tab:app_full_evaluation_experiments} gives the full E1--E6 experiment inventory, including source--target settings, matching strategies, and target mappings.

\begin{table}[!ht]
\caption{Full E1--E6 evaluation experiment inventory.
Each row summarizes one implemented perturbation family and the claim-family pairs generated from it.}
\label{tab:app_full_evaluation_experiments}
\centering
\footnotesize
\begin{tabularx}{\textwidth}{L{0.02\textwidth} L{0.1\textwidth} L{0.12\textwidth} L{0.06\textwidth} L{0.22\textwidth} Y}
\toprule
\textbf{ID} & \textbf{Perturbation} & \textbf{Source setting} & \textbf{Target setting} & \textbf{Feature comparison} & \textbf{Targets and pairs} \\
\midrule
E1
& dataset transfer
& ADNI baseline
& AIBL baseline
& FreeSurfer 8 cortical regional volume and subcortical volume, exact ROI matching
& CN--AD, CN--MCI, MMSE, and age claims are transferred from ADNI to AIBL across two volume feature spaces, giving 8 evaluated pairs. \\\hline

E2
& weak-label dataset transfer
& ADNI baseline
& OASIS-1 baseline
& FreeSurfer 8 cortical regional volume and subcortical volume, exact ROI matching
& MMSE and age claims are transferred directly, and ADNI CN--AD claims are evaluated against OASIS-1 CDR-derived impairment, giving 6 evaluated pairs. \\\hline

E3
& cortical-thickness pipeline drift
& ADNI SPM-matched
& ADNI SPM-matched
& FreeSurfer 8 Schaefer thickness versus SPM/CAT12 Schaefer thickness, with exact ROI or network-level matching depending on comparison
& CN--AD, CN--MCI, MMSE, and age claims are compared across three matched thickness comparisons, giving 12 evaluated pairs. \\\hline

E4
& volume pipeline drift
& ADNI baseline
& ADNI baseline
& FreeSurfer 8 cortical/subcortical volume versus FastSurfer cortical/subcortical volume, exact ROI matching
& CN--AD, CN--MCI, MMSE, and age claims are compared across two volume feature spaces, giving 8 evaluated pairs. \\\hline

E5
& atlas sensitivity
& ADNI baseline, AIBL baseline, OASIS-1 baseline
& same dataset view
& FreeSurfer 8 Schaefer, Kong, and Yale cortical-thickness atlases, group-level matching
& ADNI/AIBL use CN--AD, CN--MCI, MMSE, and age; OASIS-1 uses CDR-derived impairment, MMSE, and age; three atlas-pair comparisons give 33 evaluated pairs. \\\hline

E6
& feature-family sensitivity
& ADNI baseline, AIBL baseline, OASIS-1 baseline
& same dataset view
& FreeSurfer 8 cortical/subcortical volumes versus FreeSurfer 8 Schaefer cortical thickness, group-level matching
& ADNI/AIBL use CN--AD, CN--MCI, MMSE, and age; OASIS-1 uses CDR-derived impairment, MMSE, and age; two volume-to-thickness comparisons give 22 evaluated pairs. \\
\bottomrule
\end{tabularx}
\end{table}

\subsubsection{Matching strategies}
\label{app:evaluation_matching_strategies}

The evaluation configuration supports three matching regimes.
The first is exact ROI matching, which is used when source and target ROI-statistic tables share comparable feature names.
The second is group-level matching, which is used when exact ROI names do not correspond but both feature spaces can be mapped to a shared anatomical or network-level grouping.
The third, no direct ROI matching, retains metadata for comparisons whose feature spaces do not support stability computation.

Exact ROI matching defines the shared feature set as
\begin{equation}
\mathcal{J}_{st}
=
\mathcal{R}_{s_{\mathrm{src}}}
\cap
\mathcal{R}_{s_{\mathrm{tgt}}},
\end{equation}
where $\mathcal{R}_{s_{\mathrm{src}}}$ and $\mathcal{R}_{s_{\mathrm{tgt}}}$ are the source and target ROI-name sets.
Evidence, support, direction, magnitude, and transfer metrics are computed over the matched ROI set.
Exact matching is used for E1, E2, E4, and the direct Schaefer-to-Schaefer comparisons in E3.

Group-level matching defines a shared group space using an anatomical or network-level mapping.
The main group strategies are \texttt{network\_7\_like}, \texttt{lobe}, and \texttt{group\_summary}.
The fallback order prioritizes the requested grouping variable and then uses available anatomical annotations such as network, ROI group, anatomical group, lobe, or network-like group.
Group-level matching is used for atlas sensitivity and feature-family sensitivity, where one-to-one ROI correspondence is not available or would be misleading.

Table~\ref{tab:app_matching_strategies} summarizes the matching strategies used in the experiments.

\begin{table}[!ht]
\caption{Matching strategies used in E1--E6.
Exact matching is used when ROI names are comparable, while group-level matching supports atlas and feature-family comparisons.}
\label{tab:app_matching_strategies}
\centering
\scriptsize
\begin{tabularx}{\textwidth}{L{0.12\textwidth} L{0.2\textwidth} L{0.2\textwidth} Y}
\toprule
\textbf{Strategy} & \textbf{Shared unit} & \textbf{Used in} & \textbf{Purpose} \\
\midrule
\texttt{exact\_roi}
& matched ROI names
& E1, E2, E4, direct E3 comparisons
& Compares evidence and support over the same feature columns or compatible anatomical names. \\\hline

\texttt{network\_7\_ like}
& coarse network-like group
& E3 and E5 where applicable
& Compares atlas-specific cortical-thickness evidence at a shared network-like resolution. \\\hline

\texttt{lobe}
& cortical lobe or anatomical lobe group
& E5 and E6 where applicable
& Provides a coarse anatomical comparison when atlas parcels or feature families differ. \\\hline

\texttt{group\_ summary}
& ROI group or anatomical group
& E6 and fallback grouped comparisons
& Supports comparison between volumes and cortical thickness or between non-identical region systems. \\\hline

\texttt{no\_direct\_ roi\_match}
& none
& not used in the main evaluated pairs
& Records comparisons for which stability cannot be computed from the available feature spaces. \\
\bottomrule
\end{tabularx}
\end{table}

\subsubsection{Target mappings}
\label{app:evaluation_target_mappings}

Evaluation pairs may compare the same target across datasets or compare related but non-identical targets.
For direct transfer, the source and target represent the same scientific target, such as MMSE or age.
For weak-label transfer, the target definitions differ but are scientifically related.
The main weak-label case is ADNI CN--AD transferred to OASIS-1 CDR-derived impairment.

Table~\ref{tab:app_target_mappings} summarizes the target mappings used across E1--E6.

\begin{table}[!ht]
\caption{Target mappings used in evaluation experiments.
OASIS-1 is treated as a weak-label setting when CDR-derived impairment is used instead of CN/MCI/AD diagnosis.}
\label{tab:app_target_mappings}
\centering
\footnotesize
\begin{tabularx}{\textwidth}{L{0.18\textwidth} L{0.24\textwidth} L{0.24\textwidth} Y}
\toprule
\textbf{Mapping type} & \textbf{Source target} & \textbf{Target target} & \textbf{Used in} \\
\midrule
diagnosis transfer
& ADNI CN--AD
& AIBL CN--AD
& E1 \\

diagnosis transfer
& ADNI CN--MCI
& AIBL CN--MCI
& E1 \\\hline

weak-label transfer
& ADNI CN--AD
& OASIS-1 CDR-derived impairment
& E2 \\\hline

continuous transfer
& ADNI MMSE
& AIBL or OASIS-1 MMSE
& E1, E2 \\

continuous transfer
& ADNI age
& AIBL or OASIS-1 age
& E1, E2 \\\hline

within-view target preservation
& CN--AD, CN--MCI, MMSE, age
& same target in same dataset view
& E3, E4, E5, E6 for ADNI and AIBL \\

within-view weak-label preservation
& CDR-derived impairment, MMSE, age
& same target in OASIS-1 baseline
& E5, E6 for OASIS-1 \\
\bottomrule
\end{tabularx}
\end{table}

For binary contrasts, the positive and negative labels are defined in the claim-extraction setting.
For continuous targets, the target direction is retained from the source claim and the transferred claim score is evaluated by correlation with the target in the target setting.
For weak-label transfer, the target measures transfer to a related clinical construct; the resulting scores are interpreted within the target label definition.

\subsubsection{Skipped-pair and error policy}
\label{app:skipped_error_policy}

The evaluator records every configured pair as evaluated, skipped, or error.
A pair is marked as skipped when the source claim family is missing, the target claim family is missing, the source or target ROI statistics are missing, or no matched features are available under the configured matching strategy.
A pair is marked as insufficient when the comparison can be constructed but does not have enough matched units for reliable stability estimation.
A pair is marked as error only when an unexpected exception occurs during loading, matching, metric computation, transfer scoring, or label assignment.

All skipped and error cases are written to output tables with the reason.
The output tables therefore record coverage of the observed multiverse.
In the experiments reported in the main paper, the expanded evaluation configuration contained 89 pairs, all 89 pairs were evaluated, and no pair was skipped or marked as an error.

\subsection{Implementation architecture and configuration}
\label{app:implementation_details}

\subsubsection{Framework architecture}
\label{app:framework_architecture}

SCS is implemented as a configuration-driven Python package, \texttt{scs\_morph}.
The package follows the scientific workflow, with separate modules for each processing and evaluation stage.
Table~\ref{tab:app_scs_morph_architecture} summarizes the main implementation layers used in the experiments.

\begin{table}[!ht]
\caption{\texttt{scs\_morph} implementation architecture.
The framework separates data preparation, claim extraction, evaluation, and review so that each stage can be reproduced and audited independently.}
\label{tab:app_scs_morph_architecture}
\centering
\footnotesize
\begin{tabularx}{\textwidth}{L{0.15\textwidth} L{0.24\textwidth} Y}
\toprule
\textbf{Layer} & \textbf{Main modules or scripts} & \textbf{Role} \\
\midrule
data loading and audit
& \texttt{scs\_morph.data}
& Loads raw tabular inputs, audits column encodings, validates vectorized feature columns, and constructs dataset-level audit reports. \\

metadata harmonization
& \texttt{scs\_morph.harmonization}
& Harmonizes dataset-specific identifiers, demographic variables, diagnosis labels, cognitive scores, imaging metadata, and baseline or matched views. \\

feature matrix construction
& \texttt{scs\_morph.features}
& Converts vector-valued feature columns into scan-by-region matrices and aligned metadata tables for each dataset view and feature space. \\

analysis preparation
& \texttt{scs\_morph.analysis}
& Builds analysis tables, applies inclusion rules, encodes covariates, standardizes features or targets when configured, and prepares design matrices. \\

claim extraction
& \texttt{scs\_morph.claims}
& Fits claim-specific evidence models, applies support criteria, annotates regions, renders templates, and serializes claim cards and claim families. \\

claim evaluation
& \texttt{scs\_morph.evaluation}
& Constructs evaluation pairs, matches exact or grouped feature spaces, computes stability and transfer metrics, and assigns stability, transfer, and overall labels. \\

expert and LLM review
& \texttt{scs\_morph.expert}
& Selects review claims, exports review sheets, imports scored reviews, computes agreement summaries, and assigns expert-review decisions. \\
\bottomrule
\end{tabularx}
\end{table}

\subsubsection{Configuration files}
\label{app:configuration_files}

The workflow is controlled by YAML configuration files.
The files specify dataset views, feature spaces, claim-generation rules, evaluation pairs, thresholds, and review settings independently of the code.
Table~\ref{tab:app_configuration_files} lists the configuration files used by the experiments.

\begin{table}[!ht]
\caption{Configuration files used by the SCS workflow.
Configuration files define the observed multiverse, claim-generation settings, evaluation pairs, thresholds, and review protocol.}
\label{tab:app_configuration_files}
\centering
\footnotesize
\begin{tabularx}{\textwidth}{L{0.3\textwidth} Y}
\toprule
\textbf{Configuration file} & \textbf{Purpose} \\
\midrule
\texttt{paths.yaml}
& Defines raw, interim, processed, and output directories. \\

\texttt{datasets.yaml}
& Defines dataset input files and dataset identifiers. \\

\texttt{feature\_spaces.yaml}
& Defines feature-space names, source columns, feature-name files, feature families, pipelines, and atlases. \\

\texttt{claim\_generation.yaml}
& Defines claim-generation options, feature-family labels, and support settings. \\

\texttt{claim\_extraction\_all\_main.yaml}
& Defines claim-extraction settings to generate claim families and claim cards. \\

\texttt{claim\_extraction\_examples.yaml}
& Defines smaller example settings for inspection, debugging, or documentation. \\

\texttt{claim\_templates.yaml}
& Defines structured natural-language templates for claim text generation. \\

\texttt{evaluation\_pairs.yaml}
& Defines source--target evaluation pairs, perturbation types, matching strategies, targets, and transfer settings. \\

\texttt{evaluation\_thresholds.yaml}
& Defines thresholds for stability labels, transfer labels, and overall labels. \\

\texttt{expert\_review.yaml}
& Defines expert-review selection, reviewer groups, scoring columns, categorical labels, and decision rules. \\


\bottomrule
\end{tabularx}
\end{table}

The scientific workflow layers regenerate claim cards and evaluation metrics.
Manuscript-output configurations document the additional steps used to prepare paper outputs.

\subsection{Output artifacts and reproducibility workflow}
\label{app:output_artifacts_reproducibility}

\subsubsection{Major output artifacts}
\label{app:major_outputs}

The framework writes intermediate and final artifacts to structured directories.
Table~\ref{tab:app_output_artifacts} lists the artifacts produced by the workflow.
Some artifacts, such as harmonized clinical tables, scan-by-region feature matrices, and participant-level claim scores, are generated locally but are not redistributed because they contain participant-level or derived participant-level information.
Released artifacts are restricted to code, configuration files, documentation, and aggregate summaries permitted by the relevant data-use agreements.

\begin{table}[!ht]
\caption{Major output artifacts produced by the SCS workflow.
The outputs support auditing of claim generation, evaluation, and expert review.}
\label{tab:app_output_artifacts}
\centering
\footnotesize
\begin{tabularx}{\textwidth}{L{0.32\textwidth} Y}
\toprule
\textbf{Artifact} & \textbf{Content} \\
\midrule
\multicolumn{2}{l}{\textbf{Harmonized datasets and feature matrices}} \\
\midrule
\texttt{harmonized/}
& Harmonized dataset views, harmonization reports, and warnings. \\

\texttt{feature\_matrices/}
& Scan-by-region feature matrices and aligned metadata tables for each dataset view and feature space. \\

\midrule
\multicolumn{2}{l}{\textbf{Claim-card outputs: \texttt{data/processed/claim\_cards/}}} \\
\midrule
\texttt{claim\_cards.jsonl}
& Serialized claim cards generated from all analysis settings. \\

\texttt{claim\_families.jsonl}
& Serialized claim families linking all claim cards extracted from the same setting. \\

\texttt{claim\_cards\_summary.csv}
& Tabular summary of claim cards for counting, filtering, and review selection. \\

\texttt{claim\_family\_summary.csv}
& Tabular summary of claim families and source settings. \\

\texttt{per\_setting/}
& ROI-level statistics, claim-family metadata, and setting-specific artifacts. \\

\midrule
\multicolumn{2}{l}{\textbf{Evaluation outputs: \texttt{data/processed/evaluations/}}} \\
\midrule
\texttt{evaluation\_pairs\_expanded.csv}
& Expanded source--target evaluation-pair table. \\

\texttt{evaluation\_results.csv}
& One-row-per-pair evaluation results with labels and metric summaries. \\

\texttt{evidence\_stability.csv}
& Evidence-concordance metrics for evaluated pairs. \\

\texttt{support\_stability.csv}
& Support-overlap metrics for evaluated pairs. \\

\texttt{direction\_stability.csv}
& Direction-retention and sign-flip metrics. \\

\texttt{magnitude\_stability.csv}
& Magnitude-drift metrics. \\

\texttt{group\_stability.csv}
& Group-level stability metrics for non-identical feature spaces. \\

\texttt{transfer\_metrics.csv}
& Claim-score transfer metrics for binary and continuous targets. \\

\midrule
\multicolumn{2}{l}{\textbf{Expert-review outputs: \texttt{data/processed/expert\_review/}}} \\
\midrule
\texttt{selected\_main\_expert\_claims.csv}
& Main human expert review set. \\

\texttt{selected\_llm\_extended\_claims.csv}
& Extended LLM review set. \\

\texttt{selected\_reserve\_claims.csv}
& Reserve claim set for replacement or secondary review. \\

\texttt{analyzed/}
& Imported review scores, agreement summaries, human--LLM comparisons, and expert-decision tables. \\
\bottomrule
\end{tabularx}
\end{table}

\subsubsection{Reproducibility workflow}
\label{app:repository_reproducibility}

The implementation is released as a reproducible code repository.
The repository contains the \texttt{scs\_morph} package, configuration files, and command-line scripts used to regenerate the harmonized data views, feature matrices, claim cards, evaluation results, and review summaries.
Participant-level MRI data, clinical tables, scan identifiers, scan-level feature matrices, and participant-level derived outputs are not redistributed.

The end-to-end workflow is script-based.
Each step reads the configured inputs, writes explicit output artifacts, and can be rerun independently after authorized source data are placed under the configured raw-data directory.
Table~\ref{tab:app_reproducibility_workflow} summarizes the recommended execution order.

\begin{table}[!ht]
\caption{Reproducibility workflow.
The SCS evaluation can be regenerated by running the scripts in order after obtaining authorized access to the source datasets.}
\label{tab:app_reproducibility_workflow}
\centering
\begin{tabularx}{\textwidth}{L{0.03\textwidth} L{0.35\textwidth} Y}
\toprule
\textbf{Step} & \textbf{Script} & \textbf{Purpose and output} \\
\midrule
1
& \texttt{audit\_data.py}
& Audits raw input tables, summarizes column encodings, missingness, and frequent values, and writes raw-data audit reports. \\

2
& \texttt{parse\_data.py}
& Parses raw vector-valued morphometry fields and checks whether feature vectors can be interpreted consistently. \\

3
& \texttt{harmonize\_data.py}
& Harmonizes dataset-specific identifiers, visits, demographics, diagnoses, cognitive scores, imaging metadata, and dataset views. \\

4
& \texttt{build\_feature\_matrices.py}
& Converts harmonized vector-valued feature columns into scan-by-region matrices and aligned metadata tables for each configured feature space. \\

5
& \texttt{extract\_claims.py}
& Expands claim-generation settings, fits claim-extraction models, writes ROI-level statistics, and serializes claim families and claim cards. \\

6
& \texttt{evaluate\_claims.py}
& Expands source--target evaluation pairs, performs exact or group-level matching, computes stability and transfer metrics, and assigns descriptive labels. \\

7
& \texttt{report\_evaluation\_results.py}
& Summarizes evaluation outputs into experiment-level and metric-specific reports for inspection and manuscript analysis. \\

8
& \texttt{export\_expert\_review.py}
& Selects claim cards for human expert review, LLM review, and reserve review, and exports review sheets. \\

9
& \texttt{import\_expert\_reviews.py}
& Imports returned review sheets and converts reviewer scores, categorical decisions, corrected text, and comments into analysis-ready tables. \\

10
& \texttt{analyze\_expert\_reviews.py}
& Computes claim-level review summaries, agreement metrics, human--LLM agreement, high-disagreement flags, and expert-review decisions. \\
\bottomrule
\end{tabularx}
\end{table}

Workflow stages can be rerun separately.
For example, after claim cards have been generated, new evaluation pairs can be added by editing \texttt{configs/evaluation\_pairs.yaml} and rerunning only \texttt{scripts/evaluate\_claims.py}.
Similarly, new claim-review batches can be created by editing \texttt{configs/expert\_review.yaml} and rerunning the expert-review export script.
Additional datasets, pipelines, atlases, feature families, and review protocols can be incorporated through these modules while retaining the core SCS definitions.

\section{Expert and LLM Review Protocol}
\label{app:expert_review_protocol}

\subsection{Claim selection for review}
\label{app:review_selection}

SCS uses structured review to assess whether generated claim cards are scientifically plausible, clearly written, faithful to their evidence, and useful for interpretation.
The review protocol assesses claim cards containing group-level scientific statements derived from morphometry evidence.
Reviewers are instructed to evaluate these statements at the group level.

The review set is sampled from the generated claim-card pool to cover multiple datasets, feature spaces, claim levels, and claim types.
The main human expert review set contains 60 claim cards, each scored by all three human reviewers.
The extended LLM review set contains these 60 claims plus 60 additional claims, giving 120 LLM-reviewed cards in total.
A reserve set of 20 additional claims was prepared for possible replacements but was not used in the reported review.
The human review panel includes three domain perspectives: neuroscience, geriatrics, and radiology.
The LLM judges provide auxiliary review scores, with agreement to human reviewers assessed empirically.
No participant-level MRI data, scan-level feature vectors, clinical rows, subject identifiers, or derived participant-level data are provided to LLM judges; review inputs contain only aggregate claim-card text and non-identifying statistical summaries.

The selected human review set is stratified across claim levels and claim types.
It includes composite, atomic ROI, group-summary, ranking, latent-axis, latent-axis association, normative-deviation, and normative atomic ROI claims.
It also includes diagnosis contrasts, MMSE associations, group-enrichment claims, age associations, feature-ranking claims, CDR-derived impairment claims, latent morphometry-axis claims, and normative-deviation claims.
Stratification permits assessment of claim-card validity across claim granularities, targets, datasets, and feature families.

Each reviewed claim card is exported as one row in a review table.
The row contains the generated claim text, source setting, feature family, target, anatomical fields, support information, and statistical evidence required for review.
Reviewers score the claim as written, but may optionally provide a corrected claim text and short comments when the wording is misleading, too broad, insufficiently cautious, or anatomically unclear.

\subsection{Review dimensions and scoring rubric}
\label{app:review_rubric}

Each reviewer $r\in\mathcal{R}$ scores each claim $c$ on five ordinal dimensions.
For dimension $d$, the score is
\begin{equation}
a_{r,d}(c)\in\{1,2,3,4,5\}.
\label{eq:app_review_score}
\end{equation}
The dimensions are plausibility, clarity, faithfulness, usefulness, and reviewer confidence.
Table~\ref{tab:app_review_dimensions} defines the question associated with each dimension.

\begin{table}[!ht]
\caption{Expert and LLM review dimensions.
Each dimension targets a different aspect of claim-card validity.}
\label{tab:app_review_dimensions}
\centering
\begin{tabularx}{\textwidth}{L{0.21\textwidth} Y}
\toprule
\textbf{Dimension} & \textbf{Question} \\
\midrule
plausibility
& Is the claim biologically or clinically plausible given current neuroanatomical and clinical knowledge? \\

clarity
& Is the claim understandable, specific, and well phrased? \\

faithfulness
& Does the claim match the evidence shown in the row, including ROI, direction, statistics, and feature type? \\

usefulness
& Would the claim help interpret MRI morphometry results? \\

reviewer confidence
& How confident is the reviewer in their judgment of this claim? \\
\bottomrule
\end{tabularx}
\end{table}

All dimensions use the same numerical scoring rubric.
A score of 1 indicates a poor rating on the corresponding dimension, while a score of 5 indicates a strong rating.
Table~\ref{tab:app_review_score_rubric} gives the common interpretation of the 1--5 scale.

\begin{table}[!ht]
\caption{Review score rubric.
The same 1--5 scale is used for plausibility, clarity, faithfulness, usefulness, and confidence, with dimension-specific wording provided in the review sheet.}
\label{tab:app_review_score_rubric}
\centering
\begin{tabularx}{\textwidth}{L{0.06\textwidth} Y}
\toprule
\textbf{Score} & \textbf{Meaning} \\
\midrule
1
& Very low rating, such as implausible, unclear, unfaithful, not useful, or very low confidence. \\

2
& Weak rating, such as questionable, somewhat unclear, weakly faithful, low usefulness, or low confidence. \\

3
& Moderate rating, such as plausible but uncertain, understandable but improvable, mostly faithful with caution, moderately useful, or moderate confidence. \\

4
& Strong rating, such as plausible, clear, faithful, useful, or high confidence. \\

5
& Very strong rating, such as highly plausible, very clear, very faithful, highly useful, or very high confidence. \\
\bottomrule
\end{tabularx}
\end{table}

Reviewers also assign two categorical judgments.
The caution level takes values \texttt{stable}, \texttt{tentative}, \texttt{fragile}, \texttt{reject}, or \texttt{not\_sure}.
The recommended action takes values \texttt{accept}, \texttt{minor\_revise}, \texttt{major\_revise}, \texttt{reject}, or \texttt{not\_sure}.
The review instructions reward cautious, faithful wording and penalize unsupported causal, diagnostic, or mechanistic claims.
Reviewers are encouraged to use \texttt{not\_sure} and lower confidence when a claim falls outside their expertise.

\subsection{Agreement and decision rules}
\label{app:review_agreement}

SCS summarizes expert review at the claim level and at the dimension level.
For each claim $c$ and dimension $d$, the mean human score is
\begin{equation}
\bar{a}_{d}^{\mathrm{human}}(c)
=
\frac{1}{|\mathcal{R}_{\mathrm{human}}(c)|}
\sum_{r\in\mathcal{R}_{\mathrm{human}}(c)}
a_{r,d}(c),
\label{eq:app_human_mean_score}
\end{equation}
where $\mathcal{R}_{\mathrm{human}}(c)$ denotes the human reviewers who scored claim $c$.
The same summary can be computed for LLM judges or for all reviewers.

Reviewer agreement is summarized using pairwise correlations and intraclass correlation coefficients.
For each dimension $d$, SCS computes pairwise Pearson and Spearman correlations between reviewer score vectors on overlapping claims.
Let $\mathbf{a}_{r,d}$ and $\mathbf{a}_{r',d}$ denote two reviewers' score vectors for dimension $d$ over shared claims.
The pairwise Spearman agreement is
\begin{equation}
\rho_{rr',d}
=
\rho_{\mathrm{Spearman}}
\left(
\mathbf{a}_{r,d},
\mathbf{a}_{r',d}
\right).
\label{eq:app_pairwise_reviewer_correlation}
\end{equation}

For human experts, SCS also reports ICC-style agreement.
Given a claim-by-reviewer score matrix for one dimension, let $n$ be the number of claims, $k$ be the number of reviewers, $MS_R$ be the mean square for rows, $MS_C$ the mean square for columns, and $MS_E$ the residual mean square.
The single-reviewer ICC is
\begin{equation}
\mathrm{ICC}(2,1)
=
\frac{MS_R-MS_E}
{MS_R+(k-1)MS_E+\frac{k}{n}(MS_C-MS_E)}.
\label{eq:app_icc_21}
\end{equation}
The average-reviewer ICC is
\begin{equation}
\mathrm{ICC}(2,k)
=
\frac{MS_R-MS_E}
{MS_R+\frac{1}{n}(MS_C-MS_E)}.
\label{eq:app_icc_2k}
\end{equation}
These agreement summaries follow standard intraclass-correlation formulations for reviewer reliability~\citep{shrout1979intraclass,mcgraw1996forming}.

Human--LLM agreement is evaluated by comparing each LLM judge to the mean human score.
For LLM judge $\ell$ and dimension $d$, SCS computes
\begin{equation}
\rho_{\ell,d}^{\mathrm{human}}
=
\rho_{\mathrm{Spearman}}
\left(
\mathbf{a}_{\ell,d},
\bar{\mathbf{a}}_{d}^{\mathrm{human}}
\right),
\label{eq:app_human_llm_agreement}
\end{equation}
along with mean absolute difference and signed bias.
These measures quantify the agreement of auxiliary LLM judges with human experts.

SCS also assigns claim-level review decisions.
A high-disagreement flag is assigned when the human standard deviation for plausibility or faithfulness exceeds a configured threshold.
In the experiments, this threshold is 1.25.
Expert rejection is assigned when mean human plausibility or mean human faithfulness is at most 2.5, or when the majority recommended action is \texttt{reject}.
Expert acceptance or minor revision is assigned when mean human plausibility is at least 4, mean human faithfulness is at least 4, and mean human usefulness is at least 3.5, with the final label depending on the majority recommended action.
Claims that do not satisfy acceptance or rejection criteria are assigned major revision, expert disagreement, or insufficient review depending on review coverage and disagreement status.
Table~\ref{tab:app_review_decision_rules} summarizes these decision rules.

\begin{table}[!ht]
\caption{Expert-review decision rules.
Decision labels summarize human review outcomes while preserving the underlying scores, comments, caution levels, and recommended actions.}
\label{tab:app_review_decision_rules}
\centering
\begin{tabularx}{\textwidth}{L{0.23\textwidth} Y}
\toprule
\textbf{Decision} & \textbf{Rule} \\
\midrule
insufficient reviews
& Fewer than two human reviewers scored the claim. \\

expert disagreement
& Human-review standard deviation for plausibility or faithfulness is at least 1.25. \\

expert rejected
& Mean human plausibility $\leq 2.5$, mean human faithfulness $\leq 2.5$, or majority action is \texttt{reject}. \\

expert accepted
& Mean human plausibility $\geq 4$, mean human faithfulness $\geq 4$, mean human usefulness $\geq 3.5$, and majority action is \texttt{accept}. \\

expert minor revision
& Mean human plausibility $\geq 4$, mean human faithfulness $\geq 4$, mean human usefulness $\geq 3.5$, but majority action is not \texttt{accept}. \\

expert major revision
& Claim is sufficiently reviewed and not rejected, accepted, minor-revised, or high-disagreement. \\
\bottomrule
\end{tabularx}
\end{table}

Alongside statistical evaluation, the review protocol assesses whether generated claims are scientifically interpretable, appropriately cautious, and faithful to the displayed evidence.

\subsection{LLM auxiliary-review prompt}
\label{app:llm_review_prompt}

LLM judges use the same review dimensions and categorical labels as human reviewers.
The full auxiliary-review prompt is stored in \texttt{configs/llm\_expert\_review\_prompt.md}.
The prompt instructs LLM judges to act as conservative scientific reviewers, score group-level MRI morphometry claims, avoid patient-level diagnosis, avoid unsupported mechanisms, and return only a completed CSV-compatible review table.
The following excerpt documents the main review constraints.

\begin{quote}
\footnotesize
You are acting as an expert scientific reviewer for a research project titled: ``A Scientific Claim Stability Framework for Evaluating MRI Morphometry Pipelines''.

\ldots

The claims are group-level scientific statements, not individual patient diagnoses.

Important:
\begin{itemize}
    \item Do not infer patient-level diagnosis.
    \item Do not add unsupported biological mechanisms.
    \item Do not reward claims just because they sound familiar.
    \item Focus on whether the claim is plausible, clear, faithful to the provided evidence, and useful.
    \item Use conservative scientific judgment.
    \item Do not change \texttt{review\_id} or \texttt{claim\_id}.
\end{itemize}

\ldots

Please fill in the following scoring columns:
\texttt{plausibility\_1\_5}, \texttt{clarity\_1\_5}, \texttt{faithfulness\_1\_5}, \texttt{usefulness\_1\_5}, and \texttt{reviewer\_confidence\_1\_5}.

\ldots

Choose exactly one \texttt{caution\_level}: \texttt{stable}, \texttt{tentative}, \texttt{fragile}, \texttt{reject}, or \texttt{not\_sure}.

Choose exactly one \texttt{recommended\_action}: \texttt{accept}, \texttt{minor\_revise}, \texttt{major\_revise}, \texttt{reject}, or \texttt{not\_sure}.

\ldots

Return only the completed CSV table.
Preserve all original rows.
Preserve \texttt{review\_id} and \texttt{claim\_id} exactly.
Use numeric values 1--5 for score columns.
Use only the allowed categorical values.
\end{quote}

The prompt uses the same scoring structure as the human review sheet and requires CSV output, unchanged identifiers, and interpretations limited to the supplied evidence.


\section{Additional Results}
\label{app:additional_results}

\subsection{Detailed descriptions of the main figures}
\label{app:figure_descriptions}

\subsubsection{Figure~\ref{fig:scs_framework}: Scientific Claim Stability framework}
\label{app:figure_scs_framework}

\textbf{Panel A: Conventional evaluation.}
The conventional workflow converts structural MRI scans into regional morphometry tables and evaluates pipeline outputs using measurement comparisons, agreement statistics, or predictive performance.

\textbf{Panel B: Claim-centric evaluation.}
SCS extracts scientific statements from the same ROI tables and evaluates the evidence and anatomical support for each statement.
The diagram shows contrast, association, ranking or group, and latent or normative claims, alongside the complementary dimensions of stability, transfer, drift, and review.

\textbf{Panel C: Reusable \texttt{scs\_morph} framework.}
The workflow moves through data loading and audit, harmonization, feature construction, claim extraction and evaluation, systematic reporting, expert review, and review analysis.
The feedback connection shows how reviewer assessments can be used to inspect and refine generated claims.

\textbf{Panel D: Perturbation axes.}
The framework compares claims across changes in dataset, processing pipeline, atlas or ROI system, feature family, and scientific target.
The icons represent the differences captured by these axes, including sample composition, processing choices, regional representations, morphometric measurements, and target definitions.

\textbf{Panel E: Claim-card representation.}
The schematic links a structured claim object, with its type, evidence vector, support regions, weights, and metadata, to a human-readable card.
The support, stability, transfer, drift, and expert-review indicators show how evaluation results can be presented alongside a claim; the statuses shown are illustrative.

\subsubsection{Figure~\ref{fig:claim_landscape}: Claim landscape and evaluation coverage}
\label{app:figure_claim_landscape}

\textbf{Panel A: Claim-generation flow.}
The flow diagram traces the 2{,}783 generated claim cards from dataset views through cortical-thickness or volume features, grouped scientific targets, and claim levels.
Node sizes and connecting bands show how the cards are distributed across these categories, and colors identify the grouped claim types.

\textbf{Panel B: Claim-card landscape by level.}
Each horizontal bar shows the percentage composition of a claim level across diagnosis, age, cognition, confounder, and unsupervised targets.
Markers on the right give the number of cards and claim families at each level, distinguishing the large number of atomic ROI cards from the smaller groups of summary claims.

\textbf{Panel C: Dataset--feature availability.}
The left plot shows MRI data-point counts by dataset view, together with the available feature-space counts and total feature dimensions.
The adjacent matrix reports the number of claim cards generated for each dataset--feature combination; dashes indicate feature spaces that were available but not used to generate claims in the reported evaluation.

\textbf{Panel D: E1--E6 evaluation map.}
Each row reports the number of evaluated claim-family pairs in an experiment, with marker color indicating the median support Jaccard.
The E2 experiment is divided into direct cross-dataset transfer (E2a) and CDR-derived weak-label transfer (E2b).
The remaining rows cover E1 dataset transfer, E3--E4 pipeline drift, E5 atlas sensitivity, and E6 feature-family sensitivity.

\textbf{Panel E: Claim-family stability.}
The matrix cross-tabulates stability and transfer labels across the 89 evaluated pairs, with marginal totals and tags for the resulting overall-label rules.
It distinguishes stable, moderate, fragile, and insufficient stability from strong, moderate, weak, and insufficient transfer, according to the classification rules in Appendix~\ref{app:labels}.

\subsubsection{Figure~\ref{fig:stability_transfer_fragility}: Stability, transfer, and fragile support}
\label{app:figure_stability_transfer_fragility}

\textbf{Panel A: Evidence concordance versus support overlap.}
Each evaluated pair is positioned according to the Spearman concordance of its signed evidence vectors and the Jaccard overlap of its supporting regions or matched groups.
Point colors indicate the overall evaluation labels, dashed boundaries mark the stability thresholds, and crosses identify pairs for which both plotted metrics are unavailable.
This view distinguishes preservation of the global evidence ranking from preservation of the regions supporting a claim.

\textbf{Panel B: Transfer metrics.}
The upper plot shows oriented AUROC for binary-target transfer, and the lower plot shows the absolute Spearman correlation for continuous-target transfer.
Both metrics are plotted against evidence concordance, with point colors indicating the overall evaluation labels and horizontal guides marking the displayed transfer-strength regimes.

\textbf{Panel C: Pipeline drift collapse.}
The eight E4 FreeSurfer 8--FastSurfer volume pairs are connected between their evidence Spearman and support Jaccard values.
Pair identifiers and target types are shown in the inset.
Most pairs retain high evidence concordance but have much lower support overlap, illustrating the fragility observed under the primary E4 specification.

\textbf{Panel D: Fragility gap cases.}
Selected comparisons are ordered by the difference between evidence concordance and support overlap, calculated as evidence Spearman minus support Jaccard.
The adjacent markers show evidence concordance, support Jaccard, and direction retention for each case.
Large positive gaps indicate relatively preserved evidence ranking alongside low anatomical-support overlap, whereas negative gaps indicate the reverse pattern.

\textbf{Panel E: Representative claim cards.}
Six card designs illustrate composite, atomic ROI, group-summary, ranking, latent-axis, and normative-deviation claims.
Each shows how a statement can be accompanied by its source setting, anatomical support, evidence, and evaluation context.
These cards illustrate the claim formats; not every displayed claim type is included in the E1--E6 pairwise evaluations.

\subsubsection{Figure~\ref{fig:anatomy_review}: Anatomical, representation, and review summaries}
\label{app:figure_anatomy_review}

\textbf{Panel A: Lobe marker brains.}
The matrix counts supported claim cards by anatomical group and target category, including diagnosis or impairment, age association, cognition, and confounder associations.
The adjacent brain schematics show the corresponding broad support patterns for each target category.
Counts refer to generated claims assigned to anatomical categories, not to independent participants or unique anatomical regions.

\textbf{Panel B: Network and hemisphere summary.}
Mirrored horizontal bars show supported cortical-thickness claim counts by network, hemisphere, and target category.
The left and right halves distinguish the hemispheres, while the colored segments show contributions from diagnosis, age, cognition, and confounder claims.

\textbf{Panel C: Representation-sensitivity matrix.}
Rows compare the atlas and feature-family representations used in E5 and E6, while columns distinguish diagnosis or impairment, age association, cognition, and confounder targets.
Cell values report weighted mean evidence Spearman and the number of evaluated pairs; color strips summarize the corresponding stability labels.
Cells marked \texttt{n/a} indicate that no applicable evidence-concordance summary was available for that comparison.

\textbf{Panel D: Reviewer scores.}
The radar plot compares mean plausibility, clarity, faithfulness, usefulness, and confidence scores from the geriatrics, neuroscience, and radiology experts and the two auxiliary LLM reviewers.
All dimensions use the five-point review scale defined in Appendix~\ref{app:review_rubric}.

\textbf{Panel E: Human--LLM agreement.}
For each review dimension, the plot shows the Spearman correlation between an LLM reviewer's scores and the mean score assigned by the human experts across the 60 claims reviewed by both.
The annotations $b$ indicate signed scoring bias relative to the human mean, with positive values denoting higher LLM ratings.

\textbf{Panel F: Human-review decision matrix.}
The matrix cross-tabulates final human-review decisions by grouped claim type and level for the 60 human-reviewed cards.
Cell values give the number of claims in each outcome: accepted, minor revision, major revision, rejected, or expert disagreement.

\subsection{Claim landscape inventory}
\label{app:results_claim_landscape}

Tables~\ref{tab:app_claim_level_counts}--\ref{tab:app_dataset_feature_coverage} provide the numeric inventory underlying Figure~\ref{fig:claim_landscape}.
The tables summarize the generated claim-card landscape by claim level, grouped claim type, and dataset--feature coverage.
The grouped claim-type labels summarize the formal taxonomy in Appendix~\ref{app:claim_taxonomy} for visualization and reporting.

\begin{table}[!ht]
\caption{Claim-card counts by claim level.
Atomic ROI claims preserve fine-grained regional evidence, while higher-level claim types provide interpretable summaries for review and transfer.}
\label{tab:app_claim_level_counts}
\centering
\footnotesize
\begin{tabularx}{0.6\textwidth}{L{0.3\textwidth} r r}
\toprule
\textbf{Claim level} & \textbf{Claim cards} & \textbf{Proportion} \\
\midrule
\texttt{atomic\_roi} & 2{,}081 & 0.748 \\
\texttt{group\_summary} & 259 & 0.093 \\
\texttt{latent\_axis\_association} & 97 & 0.035 \\
\texttt{direction\_summary} & 84 & 0.030 \\
\texttt{composite} & 84 & 0.030 \\
\texttt{ranking} & 84 & 0.030 \\
\texttt{normative\_atomic\_roi} & 56 & 0.020 \\
\texttt{latent\_axis} & 33 & 0.012 \\
\texttt{normative\_deviation} & 5 & 0.002 \\
\midrule
\textbf{Total} & \textbf{2{,}783} & \textbf{1.000} \\
\bottomrule
\end{tabularx}
\end{table}

\begin{table}[!ht]
\caption{Claim-card counts by grouped claim type.
Grouped claim types are used for compact visualization of the claim landscape.}
\label{tab:app_claim_type_group_counts}
\centering
\footnotesize
\begin{tabularx}{0.6\textwidth}{L{0.3\textwidth} r r}
\toprule
\textbf{Grouped claim type} & \textbf{Claim cards} & \textbf{Proportion} \\
\midrule
\texttt{diagnosis\_or\_impairment} & 1{,}184 & 0.425 \\
\texttt{age} & 708 & 0.254 \\
\texttt{cognition} & 597 & 0.215 \\
\texttt{demographic\_or\_acquisition} & 261 & 0.094 \\
\texttt{latent\_or\_normative} & 33 & 0.012 \\
\midrule
\textbf{Total} & \textbf{2{,}783} & \textbf{1.000} \\
\bottomrule
\end{tabularx}
\end{table}

\begin{table}[!ht]
\caption{Dataset and feature coverage used for claim generation.
MRI rows count non-missing feature rows in the corresponding dataset view and feature group; claim-card counts are zero for resource views used for coverage accounting but not directly expanded into claim families.}
\label{tab:app_dataset_feature_coverage}
\centering
\scriptsize
\begin{tabularx}{0.94\textwidth}{L{0.16\textwidth} L{0.17\textwidth} r r r r}
\toprule
\textbf{Dataset view} & \textbf{Feature group} & \textbf{MRI rows} & \textbf{Feature spaces} & \textbf{Feature dimensions} & \textbf{Claim cards} \\
\midrule
\texttt{ADNI} & FS8 thickness & 8{,}637 & 3 & 554 & -- \\
\texttt{ADNI} & FS8 volume & 8{,}455 & 2 & 132 & -- \\
\texttt{ADNI} & FastSurfer volume & 3{,}503 & 2 & 132 & -- \\
\texttt{ADNI\_SPM\_MATCHED} & FS8 thickness & 2{,}892 & 3 & 554 & 148 \\
\texttt{ADNI\_SPM\_MATCHED} & FS8 volume & 2{,}892 & 2 & 132 & -- \\
\texttt{ADNI\_SPM\_MATCHED} & FastSurfer volume & 1{,}353 & 2 & 132 & -- \\
\texttt{ADNI\_SPM\_MATCHED} & SPM/CAT12 thickness & 2{,}892 & 2 & 404 & 300 \\
\texttt{ADNI\_baseline} & FS8 thickness & 1{,}723 & 3 & 554 & 505 \\
\texttt{ADNI\_baseline} & FS8 volume & 1{,}723 & 2 & 132 & 405 \\
\texttt{ADNI\_baseline} & FastSurfer volume & 798 & 2 & 132 & 201 \\
\texttt{AIBL} & FS8 thickness & 1{,}297 & 3 & 554 & -- \\
\texttt{AIBL} & FS8 volume & 1{,}296 & 2 & 132 & -- \\
\texttt{AIBL\_baseline} & FS8 thickness & 688 & 3 & 554 & 399 \\
\texttt{AIBL\_baseline} & FS8 volume & 685 & 2 & 132 & 261 \\
\texttt{OASIS-1} & FS8 thickness & 436 & 3 & 554 & -- \\
\texttt{OASIS-1} & FS8 volume & 436 & 2 & 132 & -- \\
\texttt{OASIS-1\_baseline} & FS8 thickness & 436 & 3 & 554 & 365 \\
\texttt{OASIS-1\_baseline} & FS8 volume & 436 & 2 & 132 & 199 \\
\bottomrule
\end{tabularx}
\end{table}

\subsection{Experiment-level evaluation summary}
\label{app:results_experiment_summary}

Table~\ref{tab:app_full_experiment_summary} expands the compact experiment-level summary in Table~\ref{tab:results_experiment_summary}.
It reports all label distributions and median metrics used to interpret the six perturbation experiments.
Direction retention is computed over the source support set.

\begin{table}[!ht]
\caption{Full E1--E6 evaluation summary.
EC denotes median evidence Spearman, SO denotes median support Jaccard, and DR denotes median direction retention over source-support regions.}
\label{tab:app_full_experiment_summary}
\centering
\scriptsize
\begin{tabularx}{\textwidth}{L{0.12\textwidth} r r L{0.08\textwidth} L{0.08\textwidth} L{0.11\textwidth} r r r r r}
\toprule
\textbf{Experiment} & \textbf{Eval.} & \textbf{Skip} & \textbf{Stability counts} & \textbf{Transfer counts} & \textbf{Overall counts} & \textbf{EC} & \textbf{SO} & \textbf{DR} & \textbf{AUROC} & \textbf{$|\rho|$} \\
\midrule
\texttt{E1\_ADNI\_TO\_ AIBL\_ VOLUME\_TRANSFER}
& 8
& 0
& stable: 6; moderate: 2
& strong: 4; moderate: 3; weak: 1
& stable-transferable: 6; transferable-drifted: 1; insufficient: 1
& 0.891
& 0.633
& 1.000
& 0.718
& 0.352 \\

\texttt{E2\_ADNI\_TO\_ OASIS\_ VOLUME\_TRANSFER}
& 6
& 0
& stable: 5; moderate: 1
& strong: 5; moderate: 1
& stable-transferable: 5; transferable-drifted: 1
& 0.842
& 0.703
& 1.000
& 0.741
& 0.634 \\

\texttt{E3\_FS8\_VS\_SPM\_ THICKNESS\_DRIFT}
& 12
& 0
& stable: 10; moderate: 2
& insufficient: 8; strong: 2; moderate: 1; weak: 1
& stable-not-transferable: 8; stable-transferable: 2; transferable-drifted: 1; insufficient: 1
& 0.875
& 1.000
& 1.000
& 0.704
& 0.363 \\

\texttt{E4\_FS8\_VS\_ FASTSURFER\_ VOLUME\_DRIFT}
& 8
& 0
& fragile: 8
& weak: 8
& fragile: 8
& 0.903
& 0.098
& 0.546
& 0.561
& 0.137 \\

\texttt{E5\_FS8\_ THICKNESS\_ATLAS\_ SENSITIVITY}
& 33
& 0
& stable: 16; moderate: 7; fragile: 10
& insufficient: 33
& stable-not-transferable: 16; fragile: 10; insufficient: 7
& 0.700
& 1.000
& 1.000
& --
& -- \\

\texttt{E6\_FS8\_VOLUME\_ VS\_THICKNESS\_ SENSITIVITY}
& 22
& 0
& stable: 4; moderate: 1; fragile: 6; insufficient: 11
& insufficient: 22
& insufficient: 12; fragile: 6; stable-not-transferable: 4
& 0.500
& 1.000
& 1.000
& --
& -- \\
\bottomrule
\end{tabularx}
\end{table}

The high median DR for E5 and E6 describes direction retention in the grouped anatomical spaces frequently used by these comparisons.
Preservation of fine-grained ROI support requires ROI-level evaluation.

\subsection{High-concordance fragile-support cases}
\label{app:results_fragile_support}

Tables~\ref{tab:app_high_concordance_fragile_cases}--\ref{tab:app_e4_fastsurfer_drift_rows} provide the pair-level metrics underlying Figure~\ref{fig:stability_transfer_fragility}.
For readability, pair identifiers are shortened to the experiment-local suffix; full \texttt{pair\_id} values are released in \texttt{figure3\_fragility\_gap\_cases.csv} and \texttt{figure3\_e4\_pipeline\_drift.csv}.
The fragility gap, defined as evidence Spearman minus support Jaccard, is used solely for visualization.

\begin{table}[!ht]
\caption{High-concordance fragile-support pairs.
These pairs satisfy evidence Spearman $\geq 0.70$ and support Jaccard $<0.20$, showing that high evidence-rank concordance can coexist with unstable support regions.}
\label{tab:app_high_concordance_fragile_cases}
\centering
\footnotesize
\begin{tabularx}{\textwidth}{L{0.06\textwidth} L{0.03\textwidth} L{0.13\textwidth} L{0.14\textwidth} L{0.08\textwidth} r r r r L{0.10\textwidth}}
\toprule
\textbf{Pair} & \textbf{Exp.} & \textbf{Source feature} & \textbf{Target feature} & \textbf{Target} & \textbf{EC} & \textbf{SO} & \textbf{DR} & \textbf{Gap} & \textbf{Overall} \\
\midrule
E4-028
& E4
& FS8 cortical volume
& FastSurfer cortical volume
& CN--MCI
& 0.896
& 0.026
& 0.188
& 0.871
& fragile \\

E4-029
& E4
& FS8 cortical volume
& FastSurfer cortical volume
& MMSE
& 0.900
& 0.053
& 0.647
& 0.847
& fragile \\

E4-027
& E4
& FS8 cortical volume
& FastSurfer cortical volume
& CN--AD
& 0.878
& 0.053
& 0.688
& 0.825
& fragile \\

E4-032
& E4
& FS8 subcortical volume
& FastSurfer subcortical volume
& CN--MCI
& 0.948
& 0.143
& 0.529
& 0.805
& fragile \\

E4-031
& E4
& FS8 subcortical volume
& FastSurfer subcortical volume
& CN--AD
& 0.926
& 0.143
& 0.563
& 0.783
& fragile \\

E4-034
& E4
& FS8 subcortical volume
& FastSurfer subcortical volume
& age
& 0.940
& 0.176
& 0.353
& 0.764
& fragile \\

E4-033
& E4
& FS8 subcortical volume
& FastSurfer subcortical volume
& MMSE
& 0.906
& 0.176
& 0.563
& 0.730
& fragile \\
\bottomrule
\end{tabularx}
\end{table}

\begin{table}[!ht]
\caption{FreeSurfer 8 versus FastSurfer row-level drift metrics.
All E4 pairs are fragile, with low support overlap across cortical and subcortical volume comparisons despite generally high evidence concordance.}
\label{tab:app_e4_fastsurfer_drift_rows}
\centering
\footnotesize
\begin{tabularx}{\textwidth}{L{0.06\textwidth} L{0.08\textwidth} L{0.08\textwidth} L{0.1\textwidth} r r r L{0.11\textwidth} L{0.06\textwidth} L{0.04\textwidth}}
\toprule
\textbf{Target} & \textbf{Feature family} & \textbf{Source feature} & \textbf{Target feature} & \textbf{EC} & \textbf{SO} & \textbf{DR} & \textbf{Transfer} & \textbf{Transfer label} & \textbf{Overall} \\
\midrule
CN--AD
& cortical volume
& FS8 cortical volume
& FastSurfer cortical volume
& 0.878
& 0.053
& 0.688
& AUROC 0.564
& weak
& fragile \\

CN--MCI
& cortical volume
& FS8 cortical volume
& FastSurfer cortical volume
& 0.896
& 0.026
& 0.188
& AUROC 0.567
& weak
& fragile \\

MMSE
& cortical volume
& FS8 cortical volume
& FastSurfer cortical volume
& 0.900
& 0.053
& 0.647
& $\rho=-0.133$
& weak
& fragile \\

age
& cortical volume
& FS8 cortical volume
& FastSurfer cortical volume
& 0.578
& 0.026
& 0.000
& $\rho=-0.169$
& weak
& fragile \\

CN--AD
& subcortical volume
& FS8 subcortical volume
& FastSurfer subcortical volume
& 0.926
& 0.143
& 0.563
& AUROC 0.548
& weak
& fragile \\

CN--MCI
& subcortical volume
& FS8 subcortical volume
& FastSurfer subcortical volume
& 0.948
& 0.143
& 0.529
& AUROC 0.558
& weak
& fragile \\

MMSE
& subcortical volume
& FS8 subcortical volume
& FastSurfer subcortical volume
& 0.906
& 0.176
& 0.563
& $\rho=-0.113$
& weak
& fragile \\

age
& subcortical volume
& FS8 subcortical volume
& FastSurfer subcortical volume
& 0.940
& 0.176
& 0.353
& $\rho=-0.141$
& weak
& fragile \\
\bottomrule
\end{tabularx}
\end{table}

The cortical-volume age pair appears in Table~\ref{tab:app_e4_fastsurfer_drift_rows}. Its evidence Spearman is below the high-concordance threshold for inclusion in Table~\ref{tab:app_high_concordance_fragile_cases}.

\subsection{E4 robustness, uncertainty, and covariate sensitivity}
\label{app:e4_robustness}

The sensitivity and uncertainty analyses use the eight E4 FreeSurfer 8--FastSurfer claim-family comparisons (cortical and subcortical volume; CN--AD, CN--MCI, MMSE, and age). Both pipelines were evaluated on the same observations, target, and design matrix within each comparison. The primary support rule combines BH-FDR-supported regions with a top-$k$ support fallback; statistical significance is assessed through BH-FDR. All results use exact ROI alignment, with 68 cortical or 64 subcortical matched ROIs and no ambiguous or duplicate matches. Literal-name and conservative one-to-one matching returned the same matched spaces.

\paragraph{Support and label sensitivity.}
Table~\ref{tab:app_e4_topk_sensitivity} varies the top-$k$ definition. Labels change for smaller support sets, whereas all eight comparisons are fragile at each tested $k=20,25,30$. Under FDR-only support at $q\leq0.05$, $0.10$, and $0.20$, none of the eight comparisons met moderate stability (some FDR supports were empty). Holding the continuous metrics fixed and varying only the reporting thresholds gave 3 stable and 5 fragile pairs under a permissive rule; the primary and more stringent rules each gave 8 fragile pairs. Categorical labels therefore depend on the declared thresholds.

\begin{table}[!ht]
\caption{E4 top-$k$ sensitivity under exact ROI matching (8 pairs per row).}
\label{tab:app_e4_topk_sensitivity}
\centering
\begin{tabular}{rrrr}
\toprule
\textbf{$k$} & \textbf{Stable} & \textbf{Moderate} & \textbf{Fragile} \\
\midrule
5 & 1 & 4 & 3 \\
10 & 0 & 3 & 5 \\
15 & 0 & 2 & 6 \\
20 (primary) & 0 & 0 & 8 \\
25 & 0 & 0 & 8 \\
30 & 0 & 0 & 8 \\
\bottomrule
\end{tabular}
\end{table}

\paragraph{Null-overlap significance.}
We tested whether the source and target supports overlapped less than random sets of their observed sizes, using an exact lower-tail hypergeometric test in each matched ROI universe ($N=68$ cortical; $N=64$ subcortical). Benjamini--Hochberg correction was applied across all eight pair-level tests, separately from the ROI-wise BH-FDR procedure used in claim extraction. Table~\ref{tab:app_e4_overlap_null} reports the exact tests for the primary hybrid support rule at $k=20$. All four cortical comparisons had significantly below-random support overlap ($q<0.05$); no subcortical comparison did. The subcortical results are compatible with chance-level overlap and provide insufficient evidence for equivalence or support preservation.

\begin{table}[!ht]
\caption{Exact lower-tail support-overlap null tests for E4 ($k=20$); $p$ is hypergeometric and $q$ is BH-adjusted across eight comparisons.}
\label{tab:app_e4_overlap_null}
\centering
\begin{tabularx}{.88\textwidth}{L{0.23\textwidth} L{0.18\textwidth} r r r r}
\toprule
\textbf{Feature} & \textbf{Target} & \textbf{Overlap} & \textbf{SO} & \textbf{$p$} & \textbf{$q$} \\
\midrule
Cortical volume & CN--AD & 2/20 & 0.053 & 0.0199 & 0.0399 \\
Cortical volume & CN--MCI & 1/20 & 0.026 & 0.0030 & 0.0121 \\
Cortical volume & MMSE & 2/20 & 0.053 & 0.0199 & 0.0399 \\
Cortical volume & Age & 1/20 & 0.026 & 0.0030 & 0.0121 \\
Subcortical volume & CN--AD & 5/20 & 0.143 & 0.3360 & 0.4480 \\
Subcortical volume & CN--MCI & 5/20 & 0.143 & 0.3360 & 0.4480 \\
Subcortical volume & MMSE & 6/20 & 0.176 & 0.5631 & 0.5631 \\
Subcortical volume & Age & 6/20 & 0.176 & 0.5631 & 0.5631 \\
\bottomrule
\end{tabularx}
\vspace{0.2em}
\parbox{\textwidth}{\textit{Note.} Overlap is the number of ROIs shared by the two 20-region support sets; SO is their Jaccard overlap. Under this null, the expected intersection is $20\times20/N$ (5.88 for cortical and 6.25 for subcortical).}
\end{table}

\paragraph{Paired bootstrap uncertainty.}
We used 300 paired bootstrap resamples per E4 comparison (2{,}400 valid resamples; zero failures), resampling identical observations for FreeSurfer 8 and FastSurfer. Table~\ref{tab:app_e4_bootstrap} reports percentile 95\% confidence intervals for evidence Spearman, support Jaccard, and source-support direction retention. Evidence concordance had lower confidence limits of 0.760--0.912 in the seven higher-concordance comparisons; cortical age was weaker. Support membership at the top-$k$ boundary can change under resampling, giving wider or discrete overlap intervals. The exact matched-size null test evaluates below-random support overlap, while bootstrap intervals describe sampling variability.

\begin{table}[!ht]
\caption{Paired bootstrap 95\% percentile intervals for E4 (300 valid resamples per pair).}
\label{tab:app_e4_bootstrap}
\centering
\scriptsize
\begin{tabularx}{\textwidth}{L{0.20\textwidth} L{0.18\textwidth} Y Y Y}
\toprule
\textbf{Feature} & \textbf{Target} & \textbf{EC 95\% CI} & \textbf{SO 95\% CI} & \textbf{DR 95\% CI} \\
\midrule
Cortical volume & CN--AD & [0.813, 0.918] & [0.000, 0.600] & [0.118, 1.000] \\
Cortical volume & CN--MCI & [0.760, 0.908] & [0.000, 0.600] & [0.000, 1.000] \\
Cortical volume & MMSE & [0.809, 0.924] & [0.000, 0.667] & [0.111, 1.000] \\
Cortical volume & Age & [0.389, 0.714] & [0.000, 0.111] & [0.000, 0.000] \\
Subcortical volume & CN--AD & [0.841, 0.961] & [0.111, 0.481] & [0.342, 1.000] \\
Subcortical volume & CN--MCI & [0.813, 0.968] & [0.053, 0.635] & [0.188, 1.000] \\
Subcortical volume & MMSE & [0.801, 0.944] & [0.111, 0.538] & [0.312, 1.000] \\
Subcortical volume & Age & [0.912, 0.957] & [0.143, 0.212] & [0.256, 0.400] \\
\bottomrule
\end{tabularx}
\end{table}

\paragraph{Acquisition and site adjustment.}
Table~\ref{tab:e4_covariate_sensitivity} separates changes due to restricting the sample to acquisition-complete observations from changes due to adjusting for field strength, cohort, and categorical SITE effects. The last three specifications use identical row sets for each pair, with common targets and source/target design matrices. The median support Jaccard increased from 0.194 under the primary covariates on acquisition-complete observations to 0.510 after adding field strength and cohort, then changed to 0.404 in the categorical site models. The site designs included 58 or 61 SITE dummy columns (59 or 62 levels, respectively). The two CN--AD site models had rank 66 of 67 with condition numbers approximately $1.28\times10^{23}$; the remaining site designs were full rank but could be ill-conditioned (condition numbers on the order of $10^8$). The site-effect models use pseudoinverse OLS. Rank deficiency in the CN--AD designs prevents identification of their individual site coefficients, while the high condition numbers in other designs require caution in interpreting those coefficients. The observed support profiles depend on the acquisition/covariate specification, which is recorded as part of each SCS analysis setting.

\subsection{Atlas and feature-family sensitivity}
\label{app:results_representation_sensitivity}

Table~\ref{tab:app_representation_sensitivity_summary} reports the representation-sensitivity values underlying Figure~\ref{fig:anatomy_review}.
The table aggregates E5 and E6 by comparison and target group.
Rows with no evaluated pairs, including demographic/acquisition targets in E5/E6, are omitted.
Weighted means use the number of evaluated pairs as weights.
Missing EC and SO values indicate comparisons where the available feature spaces did not support meaningful evidence-concordance or support-overlap computation.

\begin{table}[!ht]
\caption{Atlas and feature-family sensitivity summary.
EC denotes weighted mean evidence Spearman and SO denotes weighted mean support Jaccard; stability counts are reported as stable/moderate/fragile/insufficient.}
\label{tab:app_representation_sensitivity_summary}
\centering
\scriptsize
\begin{tabularx}{\textwidth}{L{0.04\textwidth} L{0.32\textwidth} L{0.15\textwidth} r L{0.12\textwidth} r r}
\toprule
\textbf{Exp.} & \textbf{Comparison} & \textbf{Target group} & \textbf{Pairs} & \textbf{Stability counts} & \textbf{EC} & \textbf{SO} \\
\midrule
E5
& Kong17 $\rightarrow$ Yale696
& diagnosis/impairment
& 5
& 2/2/1/0
& 0.680
& 1.000 \\

E5
& Kong17 $\rightarrow$ Yale696
& age
& 3
& 0/0/3/0
& 0.371
& 1.000 \\

E5
& Kong17 $\rightarrow$ Yale696
& cognition
& 3
& 1/1/1/0
& 0.600
& 1.000 \\

E5
& Schaefer7 $\rightarrow$ Kong17
& diagnosis/impairment
& 5
& 5/0/0/0
& 0.966
& 1.000 \\

E5
& Schaefer7 $\rightarrow$ Kong17
& age
& 3
& 2/1/0/0
& 0.771
& 1.000 \\

E5
& Schaefer7 $\rightarrow$ Kong17
& cognition
& 3
& 3/0/0/0
& 0.943
& 1.000 \\

E5
& Schaefer7 $\rightarrow$ Yale696
& diagnosis/impairment
& 5
& 2/0/3/0
& 0.520
& 1.000 \\

E5
& Schaefer7 $\rightarrow$ Yale696
& age
& 3
& 0/2/1/0
& 0.600
& 1.000 \\

E5
& Schaefer7 $\rightarrow$ Yale696
& cognition
& 3
& 1/1/1/0
& 0.500
& 1.000 \\

E6
& cortical volume $\rightarrow$ Schaefer7 thickness
& diagnosis/impairment
& 5
& 2/1/2/0
& 0.520
& 1.000 \\

E6
& cortical volume $\rightarrow$ Schaefer7 thickness
& age
& 3
& 0/0/3/0
& 0.300
& 1.000 \\

E6
& cortical volume $\rightarrow$ Schaefer7 thickness
& cognition
& 3
& 2/0/1/0
& 0.700
& 1.000 \\

E6
& subcortical volume $\rightarrow$ Schaefer7 thickness
& diagnosis/impairment
& 5
& 0/0/0/5
& --
& -- \\

E6
& subcortical volume $\rightarrow$ Schaefer7 thickness
& age
& 3
& 0/0/0/3
& --
& -- \\

E6
& subcortical volume $\rightarrow$ Schaefer7 thickness
& cognition
& 3
& 0/0/0/3
& --
& -- \\
\bottomrule
\end{tabularx}
\end{table}

The support-overlap values in Table~\ref{tab:app_representation_sensitivity_summary} measure preservation of coarse anatomical or network support under atlas or feature-family perturbation.

\subsection{Anatomical claim summaries}
\label{app:results_anatomical_summary}

These tables describe the anatomical distribution of supported claim cards generated by the claim-review pipeline.
High counts identify anatomical categories with more supported statements; their interpretation as biological localization would require vertex-wise or region-wise evidence.

Table~\ref{tab:app_lobe_target_claim_counts} summarizes supported claim counts by anatomical lobe and target group.
For each lobe--target combination, the table reports both the number of supported claim cards and the median absolute evidence value, corresponding to the summary data used for Figure~\ref{fig:anatomy_review} Panels A--B.
The largest number of supported claims appeared in temporal, frontal, parietal, and unknown/anatomically mixed regions, with temporal-lobe claims particularly frequent for diagnosis/impairment and cognition-related targets.
These counts summarize the representation of broad anatomical categories among supported claims.

\begin{table}[!ht]
\caption{Supported claim counts by lobe and target group.
The table reports the claim-card counts and median absolute evidence values underlying Figure~\ref{fig:anatomy_review} Panels A--B.}
\label{tab:app_lobe_target_claim_counts}
\centering
\footnotesize
\begin{tabularx}{0.6\textwidth}{L{0.13\textwidth} L{0.18\textwidth} r r}
\toprule
\textbf{Lobe} & \textbf{Target group} & \textbf{Claims} & \textbf{Median $|e|$} \\
\midrule
\texttt{cingulate} & age & 27 & 4.299 \\
\texttt{cingulate} & cognition & 15 & 9.540 \\
\texttt{cingulate} & demographic/acquisition & 10 & 9.104 \\
\texttt{cingulate} & diagnosis/impairment & 36 & 6.749 \\
\texttt{frontal} & age & 153 & 4.202 \\
\texttt{frontal} & cognition & 25 & 4.216 \\
\texttt{frontal} & demographic/acquisition & 32 & 92.961 \\
\texttt{frontal} & diagnosis/impairment & 135 & 4.496 \\
\texttt{insula} & age & 38 & 3.503 \\
\texttt{insula} & cognition & 3 & 10.533 \\
\texttt{insula} & demographic/acquisition & 10 & 90.718 \\
\texttt{insula} & diagnosis/impairment & 23 & 6.709 \\
\texttt{medial} & age & 2 & 3.744 \\
\texttt{medial} & diagnosis/impairment & 4 & 5.191 \\
\texttt{occipital} & age & 41 & 5.961 \\
\texttt{occipital} & cognition & 15 & 8.268 \\
\texttt{occipital} & demographic/acquisition & 14 & 4.733 \\
\texttt{occipital} & diagnosis/impairment & 42 & 5.422 \\
\texttt{other\_yale} & age & 6 & 3.309 \\
\texttt{other\_yale} & cognition & 10 & 6.044 \\
\texttt{other\_yale} & diagnosis/impairment & 12 & 5.903 \\
\texttt{parietal} & age & 37 & 4.162 \\
\texttt{parietal} & cognition & 93 & 7.933 \\
\texttt{parietal} & demographic/acquisition & 19 & 8.706 \\
\texttt{parietal} & diagnosis/impairment & 83 & 6.786 \\
\texttt{sensorimotor} & age & 18 & 3.393 \\
\texttt{sensorimotor} & cognition & 16 & 4.897 \\
\texttt{sensorimotor} & diagnosis/impairment & 17 & 4.420 \\
\texttt{temporal} & age & 120 & 6.040 \\
\texttt{temporal} & cognition & 194 & 9.203 \\
\texttt{temporal} & demographic/acquisition & 32 & 3.524 \\
\texttt{temporal} & diagnosis/impairment & 310 & 7.598 \\
\texttt{unknown} & age & 107 & 7.700 \\
\texttt{unknown} & cognition & 54 & 10.157 \\
\texttt{unknown} & demographic/acquisition & 49 & 9.199 \\
\texttt{unknown} & diagnosis/impairment & 159 & 6.787 \\
\bottomrule
\end{tabularx}
\end{table}

Table~\ref{tab:app_network_hemisphere_claim_counts} further summarizes supported cortical-thickness claim cards according to 7-network assignment, hemisphere, and target group.
This table corresponds to the network--hemisphere summary used in Figure~\ref{fig:anatomy_review} Panel B.
Across networks, Default, Control, Limbic, and Somatomotor systems contributed many supported cortical-thickness claims, although the distribution varied by hemisphere and target group.
As in the lobe-level analysis, the counts describe the distribution of claim cards; network-specific vulnerability would require a direct statistical test.

\begin{table}[!ht]
\caption{Cortical-thickness network and hemisphere claim counts.
Counts are derived from supported cortical-thickness claim cards and summarize the network--hemisphere data used in Figure~\ref{fig:anatomy_review} Panel B.}
\label{tab:app_network_hemisphere_claim_counts}
\centering
\scriptsize
\begin{tabularx}{0.60\textwidth}{L{0.12\textwidth} L{0.1\textwidth} L{0.18\textwidth} r}
\toprule
\textbf{Network} & \textbf{Hemisphere} & \textbf{Target group} & \textbf{Claims} \\
\midrule
\texttt{Cont} & LH & age & 16 \\
\texttt{Cont} & LH & cognition & 25 \\
\texttt{Cont} & LH & demographic/acquisition & 2 \\
\texttt{Cont} & LH & diagnosis/impairment & 34 \\
\texttt{Cont} & RH & age & 27 \\
\texttt{Cont} & RH & cognition & 15 \\
\texttt{Cont} & RH & demographic/acquisition & 7 \\
\texttt{Cont} & RH & diagnosis/impairment & 31 \\
\texttt{Default} & LH & age & 31 \\
\texttt{Default} & LH & cognition & 59 \\
\texttt{Default} & LH & demographic/acquisition & 8 \\
\texttt{Default} & LH & diagnosis/impairment & 93 \\
\texttt{Default} & RH & age & 28 \\
\texttt{Default} & RH & cognition & 36 \\
\texttt{Default} & RH & demographic/acquisition & 8 \\
\texttt{Default} & RH & diagnosis/impairment & 61 \\
\texttt{DorsAttn} & LH & age & 6 \\
\texttt{DorsAttn} & LH & cognition & 15 \\
\texttt{DorsAttn} & LH & demographic/acquisition & 4 \\
\texttt{DorsAttn} & LH & diagnosis/impairment & 5 \\
\texttt{DorsAttn} & RH & age & 1 \\
\texttt{DorsAttn} & RH & cognition & 5 \\
\texttt{DorsAttn} & RH & demographic/acquisition & 4 \\
\texttt{DorsAttn} & RH & diagnosis/impairment & 6 \\
\texttt{Limbic} & LH & age & 12 \\
\texttt{Limbic} & LH & cognition & 15 \\
\texttt{Limbic} & LH & demographic/acquisition & 4 \\
\texttt{Limbic} & LH & diagnosis/impairment & 34 \\
\texttt{Limbic} & RH & age & 5 \\
\texttt{Limbic} & RH & cognition & 9 \\
\texttt{Limbic} & RH & demographic/acquisition & 3 \\
\texttt{Limbic} & RH & diagnosis/impairment & 25 \\
\texttt{SalVentAttn} & LH & age & 15 \\
\texttt{SalVentAttn} & LH & cognition & 5 \\
\texttt{SalVentAttn} & LH & demographic/acquisition & 6 \\
\texttt{SalVentAttn} & LH & diagnosis/impairment & 19 \\
\texttt{SalVentAttn} & RH & age & 19 \\
\texttt{SalVentAttn} & RH & cognition & 5 \\
\texttt{SalVentAttn} & RH & demographic/acquisition & 5 \\
\texttt{SalVentAttn} & RH & diagnosis/impairment & 19 \\
\texttt{SomMot} & LH & age & 33 \\
\texttt{SomMot} & LH & demographic/acquisition & 4 \\
\texttt{SomMot} & LH & diagnosis/impairment & 22 \\
\texttt{SomMot} & RH & age & 31 \\
\texttt{SomMot} & RH & demographic/acquisition & 2 \\
\texttt{SomMot} & RH & diagnosis/impairment & 24 \\
\texttt{Vis} & LH & age & 14 \\
\texttt{Vis} & LH & cognition & 6 \\
\texttt{Vis} & LH & demographic/acquisition & 14 \\
\texttt{Vis} & LH & diagnosis/impairment & 20 \\
\texttt{Vis} & RH & age & 17 \\
\texttt{Vis} & RH & cognition & 5 \\
\texttt{Vis} & RH & demographic/acquisition & 17 \\
\texttt{Vis} & RH & diagnosis/impairment & 17 \\
\bottomrule
\end{tabularx}
\end{table}

\subsection{Expert and LLM review outcomes}
\label{app:results_review_summary}

Tables~\ref{tab:app_human_expert_decisions}--\ref{tab:app_human_llm_agreement} provide the numerical review summaries underlying Figure~\ref{fig:anatomy_review} Panels D--F.
The tables report final human expert decisions and reviewer scores for claim plausibility, clarity, faithfulness, usefulness, and reviewer confidence.
Rows with \texttt{insufficient\_reviews} identify LLM-only extended claims without complete human review.

Table~\ref{tab:app_human_expert_decisions} reports the final decision distribution for the 60 claims with complete human expert review.
Most reviewed claims were judged to require either major or minor revision, while a smaller subset was accepted without revision or rejected.
The review indicates that generated claims were often scientifically useful and required expert correction, refinement, or qualification before publication.

\begin{table}[!ht]
\caption{Human expert decision counts.
Counts include only claims with complete human expert review.}
\label{tab:app_human_expert_decisions}
\centering
\scriptsize
\begin{tabularx}{0.35\textwidth}{L{0.25\textwidth} r}
\toprule
\textbf{Expert decision} & \textbf{Claims} \\
\midrule
\texttt{expert\_major\_revision} & 19 \\
\texttt{expert\_minor\_revision} & 17 \\
\texttt{expert\_accepted} & 15 \\
\texttt{expert\_rejected} & 5 \\
\texttt{expert\_disagreement} & 4 \\
\midrule
\textbf{Total human-reviewed claims} & \textbf{60} \\
\bottomrule
\end{tabularx}
\end{table}

\newpage
Table~\ref{tab:app_review_scores_by_reviewer} summarizes the mean 1--5 scores assigned by each reviewer across the five review dimensions.
The LLM reviewers tended to assign high clarity scores, whereas human experts showed more variation across domains and generally provided more conservative assessments.
These results support auxiliary LLM review for large-scale screening, with human expert review as the reference for scientific judgment.

\begin{table}[!ht]
\caption{Mean review scores by reviewer and dimension.
Scores use a 1--5 scale, with LLM reviewers providing auxiliary assessments alongside human experts.}
\label{tab:app_review_scores_by_reviewer}
\centering
\scriptsize
\begin{tabularx}{0.7\textwidth}{L{0.2\textwidth} L{0.06\textwidth} r r r r r}
\toprule
\textbf{Reviewer} & \textbf{Type} & \textbf{Plaus.} & \textbf{Clarity} & \textbf{Faith.} & \textbf{Useful.} & \textbf{Conf.} \\
\midrule
\texttt{chatgpt\_5\_5\_thinking} & LLM & 3.875 & 4.783 & 4.525 & 3.625 & 4.175 \\
\texttt{gemini\_3\_1\_pro} & LLM & 4.225 & 4.950 & 3.942 & 4.408 & 3.892 \\
\texttt{geriatrics} & human & 4.150 & 3.617 & 3.750 & 3.967 & 3.517 \\
\texttt{neuroscience} & human & 3.683 & 2.883 & 3.550 & 3.500 & 2.767 \\
\texttt{radiology} & human & 4.300 & 3.467 & 4.100 & 4.117 & 4.533 \\
\bottomrule
\end{tabularx}
\end{table}

Table~\ref{tab:app_human_review_agreement} reports agreement among the three human experts.
Agreement was highest for plausibility and clarity, moderate for faithfulness and usefulness, and lowest for reviewer confidence.
The comparison of ICC$(2,1)$ with ICC$(2,k)$ indicates higher reliability for average scores across human reviewers than for individual scores.

\begin{table}[!ht]
\caption{Human agreement by review dimension.
ICC$(2,1)$ estimates single-reviewer reliability, while ICC$(2,k)$ estimates reliability of the average human-reviewer score.}
\label{tab:app_human_review_agreement}
\centering
\scriptsize
\begin{tabularx}{0.87\textwidth}{L{0.22\textwidth} r r r r r r}
\toprule
\textbf{Dimension} & \textbf{Claims} & \textbf{Reviewers} & \textbf{ICC$(2,1)$} & \textbf{ICC$(2,k)$} & \textbf{Pearson} & \textbf{Spearman} \\
\midrule
\texttt{plausibility\_1\_5} & 60 & 3 & 0.426 & 0.690 & 0.475 & 0.400 \\
\texttt{clarity\_1\_5} & 60 & 3 & 0.398 & 0.665 & 0.525 & 0.506 \\
\texttt{faithfulness\_1\_5} & 60 & 3 & 0.360 & 0.628 & 0.379 & 0.393 \\
\texttt{usefulness\_1\_5} & 60 & 3 & 0.291 & 0.552 & 0.292 & 0.254 \\
\texttt{reviewer\_confidence\_1\_5} & 60 & 3 & 0.106 & 0.263 & 0.296 & 0.295 \\
\bottomrule
\end{tabularx}
\end{table}

Table~\ref{tab:app_human_llm_agreement} compares each LLM reviewer with the mean human expert score on the 60 fully reviewed claims.
The strongest LLM--human rank agreement was observed for ChatGPT on plausibility, while agreement was weaker for clarity and several Gemini dimensions.
Both LLM reviewers showed positive bias for clarity, rating claim wording more favorably than human experts.

\begin{table}[!ht]
\caption{Human--LLM agreement by review dimension.
Agreement is measured between each LLM reviewer and the mean human expert score on the 60 claims with complete human review.}
\label{tab:app_human_llm_agreement}
\centering
\scriptsize
\begin{tabularx}{0.83\textwidth}{L{0.2\textwidth} L{0.2\textwidth} r r r r}
\toprule
\textbf{LLM reviewer} & \textbf{Dimension} & \textbf{Overlap} & \textbf{Spearman} & \textbf{MAD} & \textbf{Bias} \\
\midrule
\texttt{chatgpt\_5\_5\_thinking} & \texttt{plausibility\_1\_5} & 60 & 0.670 & 0.594 & -0.128 \\
\texttt{chatgpt\_5\_5\_thinking} & \texttt{clarity\_1\_5} & 60 & 0.292 & 1.378 & 1.278 \\
\texttt{chatgpt\_5\_5\_thinking} & \texttt{faithfulness\_1\_5} & 60 & 0.429 & 0.889 & 0.700 \\
\texttt{chatgpt\_5\_5\_thinking} & \texttt{usefulness\_1\_5} & 60 & 0.519 & 0.778 & -0.178 \\
\texttt{chatgpt\_5\_5\_thinking} & \texttt{reviewer\_confidence\_1\_5} & 60 & 0.322 & 0.733 & 0.611 \\
\midrule
\texttt{gemini\_3\_1\_pro} & \texttt{plausibility\_1\_5} & 60 & 0.439 & 0.706 & 0.006 \\
\texttt{gemini\_3\_1\_pro} & \texttt{clarity\_1\_5} & 60 & 0.161 & 1.589 & 1.578 \\
\texttt{gemini\_3\_1\_pro} & \texttt{faithfulness\_1\_5} & 60 & 0.163 & 1.061 & 0.317 \\
\texttt{gemini\_3\_1\_pro} & \texttt{usefulness\_1\_5} & 60 & 0.163 & 0.928 & 0.572 \\
\texttt{gemini\_3\_1\_pro} & \texttt{reviewer\_confidence\_1\_5} & 60 & 0.351 & 0.689 & 0.278 \\
\bottomrule
\end{tabularx}
\end{table}



\end{document}